\documentclass[11pt,twoside,titlepage]{cslarticle}
\makeatletter
\renewenvironment{abstract}{
      \if@twocolumn
        \section*{\abstractname}
      \else
        \small
        \begin{center}
          {\bfseries \abstractname\vspace{-.5em}\vspace{\z@}}
        \end{center}
        \quotation
      \fi}
      {\if@twocolumn\else\endquotation\fi}
\makeatother
\def\telephonenum{(650) 859-2000}
\def\faxphonenum{that's last century}
\cslreportnumber{CSL Technical Report}
\acknowledge{This research was sponsored by SRI International and by
my retirement plan}
\usepackage{fancyhdr}
\usepackage{xcolor}
\usepackage{graphicx,cite,url,relative,alltt}
\usepackage[bookmarks=true,hyperfigures=true,colorlinks=true,linkcolor=blue,citecolor=blue,pagebackref=true,pdfpagemode=fullscreen,plainpages=false,pdfpagelabels]{hyperref}

\def\BigLaTeX{{\rm L\kern-.36em\raise.3ex\hbox{\smaller\smaller A}\kern-.15em
    T\kern-.1667em\lower.7ex\hbox{E}\kern-.125emX}}
\def\BoldLaTeX{{\bf L\kern-.36em\raise.3ex\hbox{\smaller\smaller\bf A}\kern-.15em
    T\kern-.1667em\lower.7ex\hbox{E}\kern-.125emX}}
\def\BibTeX{{\rm B\kern-.05em{\sc i\kern-.025em b}\kern-.08em
    T\kern-.1667em\lower.7ex\hbox{E}\kern-.125emX}}

\newlength{\hsbw}
\def\extrawidth{0.5in}

\newcounter{sessioncount}
\newenvironment{session*}{\begin{flushleft}
 \refstepcounter{sessioncount}
 \setlength{\hsbw}{\linewidth}
 \addtolength{\hsbw}{-\arrayrulewidth}
 \addtolength{\hsbw}{-\tabcolsep}
 \begin{tabular}{@{}|c@{}|@{}}\hline 
 \begin{minipage}[b]{\hsbw}
 \vspace*{-.5pt}
 \begin{flushright}
 \rule{0.01in}{.15in}\rule{0.3in}{0.01in}\hspace{-0.35in}
 \raisebox{0.04in}{\makebox[0.3in][c]{\footnotesize \thesessioncount}}
 \end{flushright}
 \vspace*{-.57in}
 \begingroup\small\vspace*{1.0ex}\begin{alltt}}{\end{alltt}\endgroup\end{minipage}\\ \hline 
 \end{tabular}
 \end{flushleft}}
\def\sessionsize{\small}
\def\smallsessionsize{\small}

\newcommand{\exmemo}[1]{}

\newcommand{\comment}[1]{}

\newcommand{\exfootnote}[1]{}
\newlength{\sblen}
\newlength{\overhang}

\def\SetFigFont#1#2#3{\rm}

\newcommand{\excite}[1]{}

\def\srilogo{\includegraphics[height=18mm]{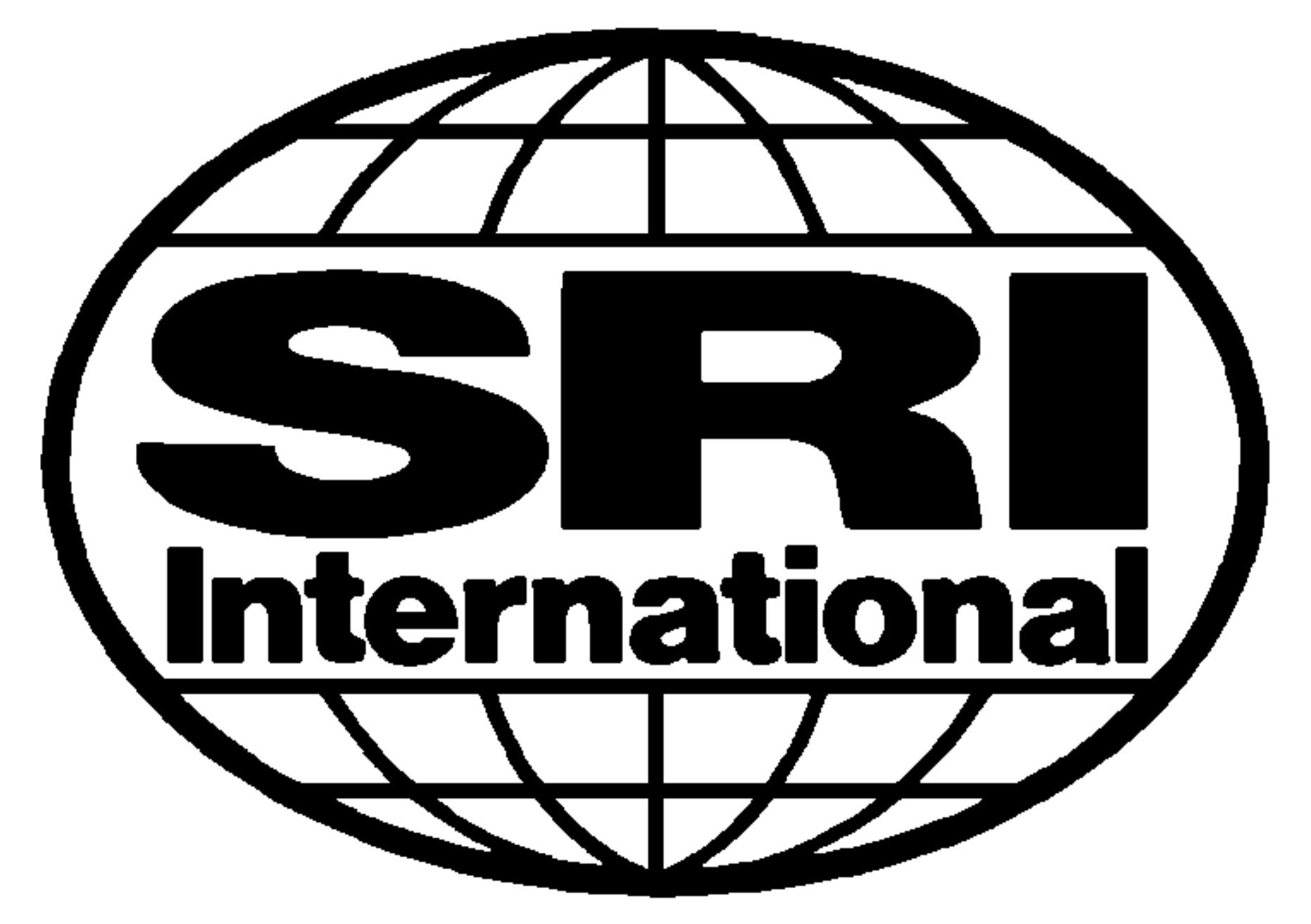}}

\newcommand{\arxiv}[1]{\href{https://arxiv.org/abs/#1}{\tt arXiv:#1}}

\usepackage{nobibhead}
\title{On Computational Mechanisms for Shared Intentionality\\[0.3ex]And Speculation on Rationality and Consciousness}
\author{John Rushby\\
\emph{Computer Science Laboratory}\\
\emph{SRI International, Menlo Park CA 94025 USA}
}
\begin{document}
\maketitle
\clearpage
\thispagestyle{empty}
\tableofcontents

\listoffigures
\clearpage
\setcounter{page}{1}
\begin{center}
\textbf{\Large On Computational\,Mechanisms for Shared\,Intentionality\\[0.5ex]And Speculation on Rationality and Consciousness}\\[0.25in]
  John Rushby\\
  \emph{Computer Science Laboratory}\\
  \emph{SRI International, Menlo Park CA 94025 USA}\\
  \texttt{Rushby@csl.sri.com}
\end{center}

\vspace*{0.25in}

\begin{abstract}

A singular attribute of humankind is our ability to undertake novel,
cooperative behavior, or teamwork.  This requires that we can
communicate goals, plans, and ideas between the brains of individuals
to create \emph{shared intentionality}.  I adopt the view that the
brain performs computation, then, using the information processing
model of David Marr, I derive necessary characteristics of basic
mechanisms to enable shared intentionality between prelinguistic
computational agents.

More speculatively, I suggest the mechanisms derived by this thought
experiment apply to humans and extend to provide explanations for
human rationality and aspects of intentional and phenomenal
consciousness that accord with observation.  This yields what I call
the Shared Intentionality First Theory (SIFT) for language,
rationality and consciousness.

The significance of shared intentionality has been recognized and
advocated previously, but typically from a sociological or behavioral
point of view.  SIFT complements prior work by applying a computer
science perspective to the underlying mechanisms.

\end{abstract}

\thispagestyle{fancy}
\section{Introduction}

\begin{figure}[ht]
\includegraphics[width=\textwidth]{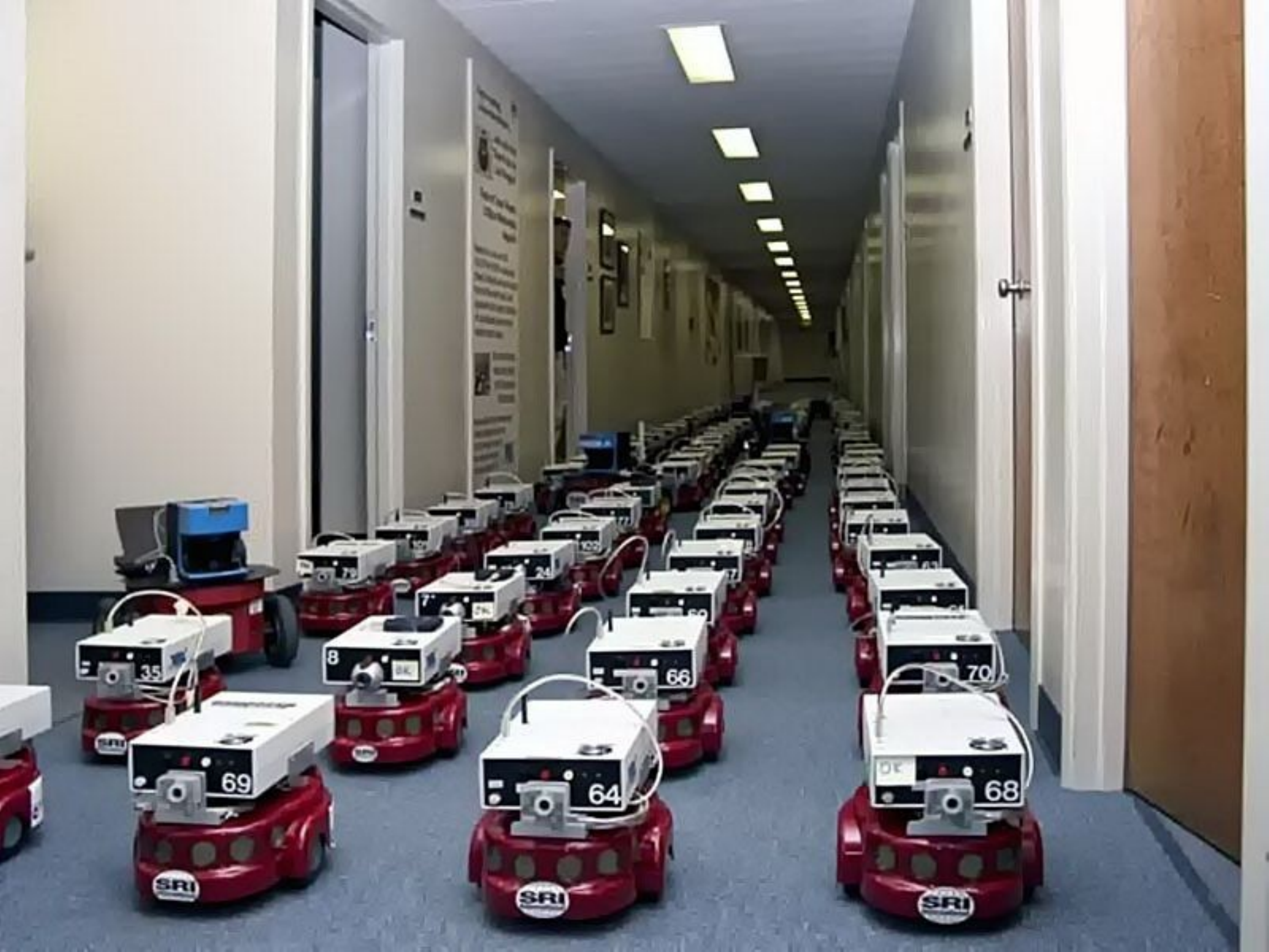}
\centering{\url{https://commons.wikimedia.org/wiki/File:SRI_Robotics_Centibots.png}\\\textcolor{magenta}{Creative Commons Attribution-Share Alike 3.0 Unported License}}
\caption{\label{Centibots}Centibots}
\end{figure}

About 20 years ago, our neighboring AI laboratory engaged in a project
named \emph{Centibots} \cite{Konolige-Centibots06}.  The Centibots
were small mobile robots, each about a foot cube, and there were a
hundred of them, hence the name (see Figure \ref{Centibots}).  They
were equipped with cameras and other sensors, and had a limited
ability to communicate with each other.

The idea was that Centibots would be deposited in a building and would
then spread out and collaboratively develop a map of the internal
layout.  To do this, each Centibot would exchange some state
information with others in its vicinity and apply this, together with
information from its own sensors, to develop and execute a continually
updated plan to achieve its part of the overall goal.  Observe that
the function and behavior of the Centibots was not unlike that of
social animals, such as bees and wolves.

Let us now leave the historical Centibots, and conduct a thought
experiment.  Suppose we took one of the Centibots and reprogrammed it
with some additional planning capabilities and a new goal: to form,
together with its cohorts, a line that can guide humans to an exit in
case of emergency.  It is obvious that no progress can be made in
this endeavor unless there is some way to share the new goal with the
other Centibots.  But the standard Centibots do not have this
capability: their communications ``language'' is limited to the
information needed for their original task, rather like the ``dances''
of bees and the howls and body postures of wolves.  Furthermore,
sharing the goal might not be enough: the standard Centibot planner
might not be able to come up with local actions to achieve the new goal,
so it may be necessary for the modified Centibot also to communicate
some ``hints'' or ``ideas'' to augment the local planners.
This cannot be accomplished by simply watching the modified Centibot
executing part of the goal (e.g., standing by an exit).

We see that to get from preprogrammed collaborative behavior to the
ability to jointly undertake \emph{new} tasks requires additional
capabilities that allow novel information to be communicated from one
Centibot to another in such a way that the recipient comes to share
some of the goals and planning elements of the sender.

Just as the basic Centibots can serve as crude models for social
insects and animals, so the hypothetically augmented Centibots can
serve as models for animals that can engage in novel forms of
collaborative behavior: that is, in teamwork.  The additional
capabilities of the augmented Centibots are an abstracted
characterization of what, in animals, is termed ``shared
intentionality'': that is, the ability to communicate and share
similar mental states that can drive similar behavior
\cite{Tomasello&Carpenter07}.\footnote{Intentionality is the property
of computational or mental states being \emph{about} or \emph{directed
toward} something but we (or computational agents) can have many different mental
attitudes toward that something besides intentions: we may, for
example, believe it, fear it, prefer it to something else, or want it,
and so on.  The philosopher's jargon ``intentional'' (due to Franz
Brentano) comes by way of translation from German and should not be
construed to refer specifically to ``intentions.''}

I believe that modern humans are the only animals that engage in
full-fledged teamwork\footnote{For example, ``it is inconceivable that you
would ever see two chimpanzees carrying a log together'' \cite[quoting
Tomasello on page 238]{Haidt13}.} and this (and its larger
manifestation as culture \cite{Searle:social-world10}) is the reason
we dominate the world.  Furthermore, I propose that our other defining
capabilities---language, rationality, consciousness---arose from the
mechanisms that enable teamwork: that is, from shared intentionality.
I will develop a computer science description of how shared
intentionality might be constructed in hypothesized post-Centibot
computational agents.  I will then argue that these mechanisms extend
to real humans\footnote{I take it as a given that the function of the
brain is to perform computations.  This is not a metaphor or simile, as when in
earlier times the brain was said to be ``like'' a clock; we are saying
it \emph{is} a computer.  This confuses some who are familiar only
with desktop computers: the brain is a computer in the sense that it
performs computations, but it is obviously organized, implemented, and
deployed completely differently than a desktop computer; for a
mechanistic analogy, think of the computational system of a
self-driving car and its integration with perception and actuation.}
and provide, almost immediately, the capabilities for rational
deliberation; I speculate that awareness of these processes
constitutes intentional consciousness.  Phenomenal consciousness
arises because we then need the ability to communicate the content of
our sense experience.

I refer to this collection of computer science, deduction, and
speculation as the ``Shared Intentionality First'' Theory of
rationality and consciousness (SIFT) and I argue that it explains
otherwise puzzling aspects of these faculties and is evolutionarily
plausible.  SIFT is more abstract than other theories of
consciousness: it concerns the purpose, strategy, and architecture of
mental mechanisms, not their biological implementation, and so it is
compatible with, or provides context for, several other theories of
brain organization and consciousness, including predictive processing
\cite{Clark13,Hohwy:book13}, higher-order thought
\cite{Gennaro04,Rosenthal05} and its related notion of metacognition
\cite{Brown-etal19}, dual-process theories
\cite{Frankish10,Kahneman11}, and global workspace theories
\cite{Baars05:GWT,Dehaene14}.  SIFT also brings a different
perspective to prior work on the r{\^o}le of shared intentionality in
the development of human cognition and language
\cite{Tomasello:origins-communication10,Tomasello:origins-cognition08,Bickerton:species90,Bickerton:Adam09}:
in particular, SIFT's computer science focus on the underlying
mechanisms complements prior work on the behavioral and psychological
attributes of shared intentionality and suggests how these form an
integrated cognitive ``package.''

The paper is organized as follows: the next section considers the
architecture of mechanisms for construction of shared intentionality
in humanoid computational agents.  Although this description is driven
by intuitions, mechanisms, and technologies from computer science and
computational models of perception, it is written at a tutorial level
and is intended to be accessible to all.  I then
propose that real humans have similar mechanisms; this is followed by
sections on rationality, intentional consciousness, and phenomenal
consciousness, respectively.  I then briefly consider the
archaeological record and the evolutionary plausibility of SIFT,
followed by comparison with some other theories and with attempts to
create artificial consciousness and conclude with a summary of
fundamental claims and acknowledgments.

\section{The Construction of Shared Intentionality}

I will couch my construction of shared intentionality in terms of
hypothesized human-like computational agents (i.e., mobile robots)
that operate in the open world and could be considered humanoid
``descendants'' of Centibots: by ``humanoid'' I mean their sensors and
actuators resemble human senses and limbs, their sense
interpretation, planning, and execution capabilities are powerful and
comparable to those of the human subconscious, and they are autonomous
(i.e., set their own goals),\footnote{I recognize the difficulties
here: what exactly are the capabilities of the human subconscious, and
how can we have autonomy without encountering the philosophical
problems of free will?  However, for the purposes required here, I
believe these topics can be set aside (i.e., we can suspend disbelief)
and that some approximate interpretation is sufficient.} but they lack
mechanisms for novel collaborative behavior: no language or other
means to exchange state information or ideas beyond those
preprogrammed for specific tasks as in the original Centibots.  I
make these choices because I will later argue that these agents, when
equipped with shared intentionality, serve as models for real humans,
and for this reason I will sometimes use anthropomorphic terms in my
discussion.

\begin{figure}[h!t]
\centering{\includegraphics[width=0.85\textwidth]{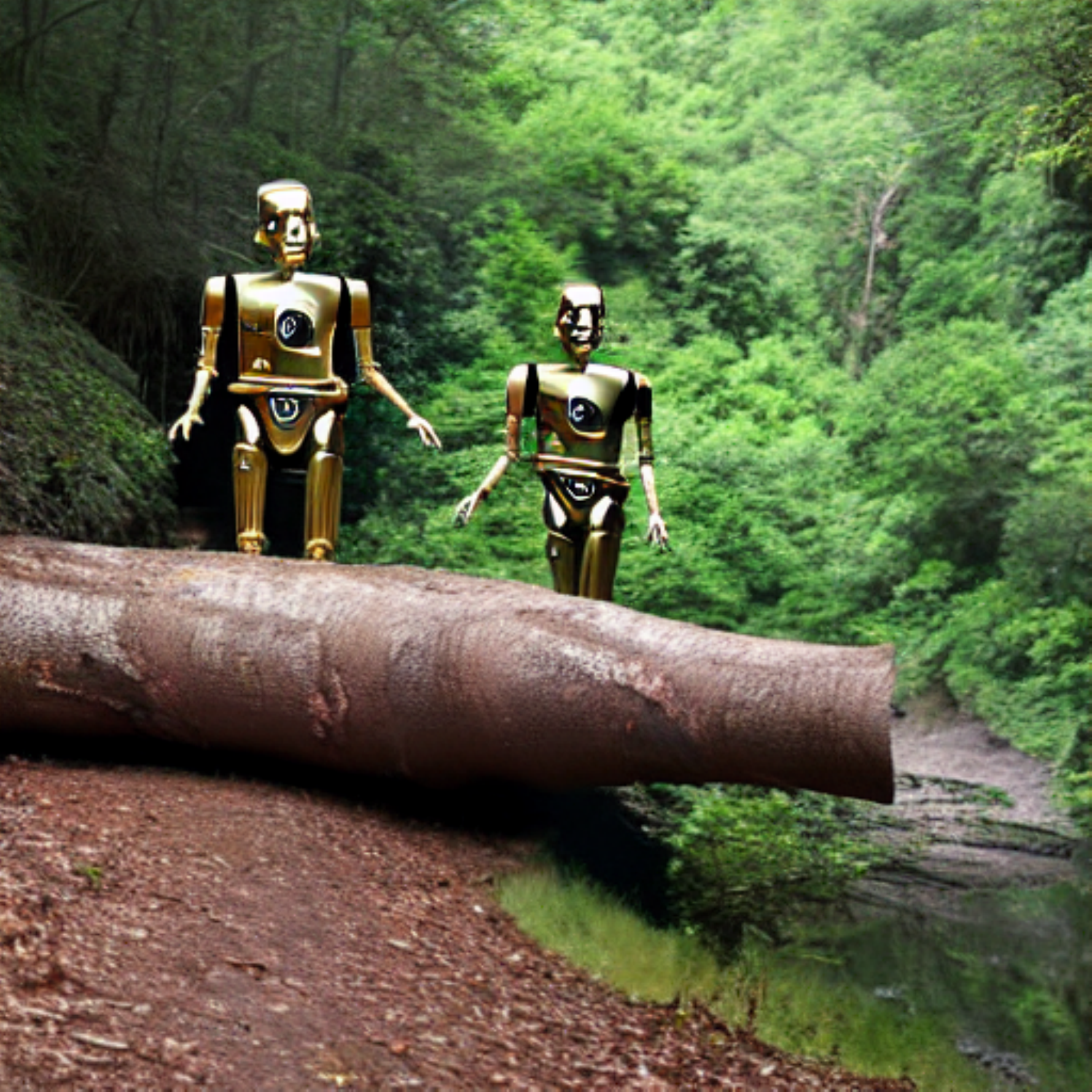}}
\\\textcolor{magenta}{Image created by DreamStudio, CC0 1.0 Universal
Public Domain Dedication}
\caption{\label{robots}Two Humanoid Agents and a Log-Bridge}
\end{figure}

In particular, let us imagine an individual agent of this type faced
with a problem: crossing a small ravine.  Thanks to its large
modeling, search, and planning capabilities our individual---let's
call it/her Alice---might invent a novel behavior: use a fallen tree
trunk or log to create a bridge across the ravine.  Now let us suppose
the log is too large and heavy for Alice to move into place.  A second
agent---let's call it/him Bob\footnote{Alice and Bob are a standard
trope for describing distributed algorithms in computer science (see
Wikipedia \cite{alice-bob}).  Their appearance here does not indicate
any specific debt to other papers that happen to have adopted the same
usage.}---may watch her struggling with the log but, having no prior
experience of log-bridges, he will not understand what is going on any
more than would a dog or a human infant, and will render no useful
assistance (see Figure \ref{robots}).\footnote{It is possible that Bob
will copy Alice's activity as a preprogrammed form of cooperation.
However, not understanding the purpose, he might very likely push the
log in the wrong direction, or push a different log.}

Notice that the behavior required in cooperative construction of a
log-bridge is quite different than simply copying from observation.
If Alice has a technique for using stones to crack nuts, Bob may be
able to learn this by simply observing and copying her behavior.
Alice does not need to explicitly communicate the skill to Bob
(indeed, she need not be aware that she has the skill, as seems to be
the case with chimpanzees), and Bob does not need to infer the goal
and technique: they are there before him.  But the log-bridge exists
only in Alice's head: Bob must somehow infer this goal and Alice must
help him do so, and likewise the plan to achieve it by moving a
specific log into place.  To get truly cooperative behavior on a novel
task, the individuals must have ``shared intentionality'': that is,
similar computational or mental states that can drive similar behavior
\cite{Tomasello&Carpenter07}.

In order to examine mechanisms that could bring this about, I adopt
and adapt the three-stage ``information processing'' model of David
Marr \cite{Marr82}.  The model concerns human (or animal) cognition
but presupposes that mental operations are information processing
tasks (i.e., computations), so it applies directly to our hypothesized
computational agents.  The first step or stage in understanding such a
computation is to deduce its goal and the strategy by which it can be
carried out.  The second stage considers the representations that can
be employed for the inputs and outputs of the computation and the
mechanism or ``algorithm'' that can transform the one into the other.
The third and final stage considers the implementation that can
physically realize the representation and algorithm: digital circuits
and software in the case of our agents, and neurons and other biology
in the case of humans and animals.

The goal of the computation under consideration is the creation of
shared intentionality and I already staked out a position when I
suggested this requires the parties to achieve ``similar computational
or mental states that can drive similar behavior.''  Thus, I will
suppose that the immediate goal is to recreate in Bob some aspects of
Alice's computational state (specifically, that associated with the
``idea'' of a log-bridge), so that his planner has access to similar
information and may use this to generate usefully cooperative behavior
\cite{Bratman13}.  At this point, we must adjust Marr's model a little
because the computation underlying shared intentionality must surely
be a \emph{distributed} one: some of it will be performed by Alice and
some by Bob, and something will be transferred between them.  We now
need to consider what are the separate computations, and what is
transferred.

Alice and Bob might be constructed differently (e.g., Alice might be
programmed in Java and Bob in Python) so we cannot simply copy some
bag of bytes from Alice's state to Bob's and expect it to have a
useful effect.  Similarly, if Alice and Bob were biological entities,
their brains, even as conspecifics, will have grown and developed
somewhat differently due to their individual genetics, physiology, and
experiences, and so the transfer mechanism cannot directly reconstruct
part of Alice's neural state (i.e., the electrical and chemical
activity in some specific cluster of neurons) in Bob's brain since
they will be ``wired up'' differently at the neural
level.

\begin{figure}[t]
\includegraphics[width=\textwidth]{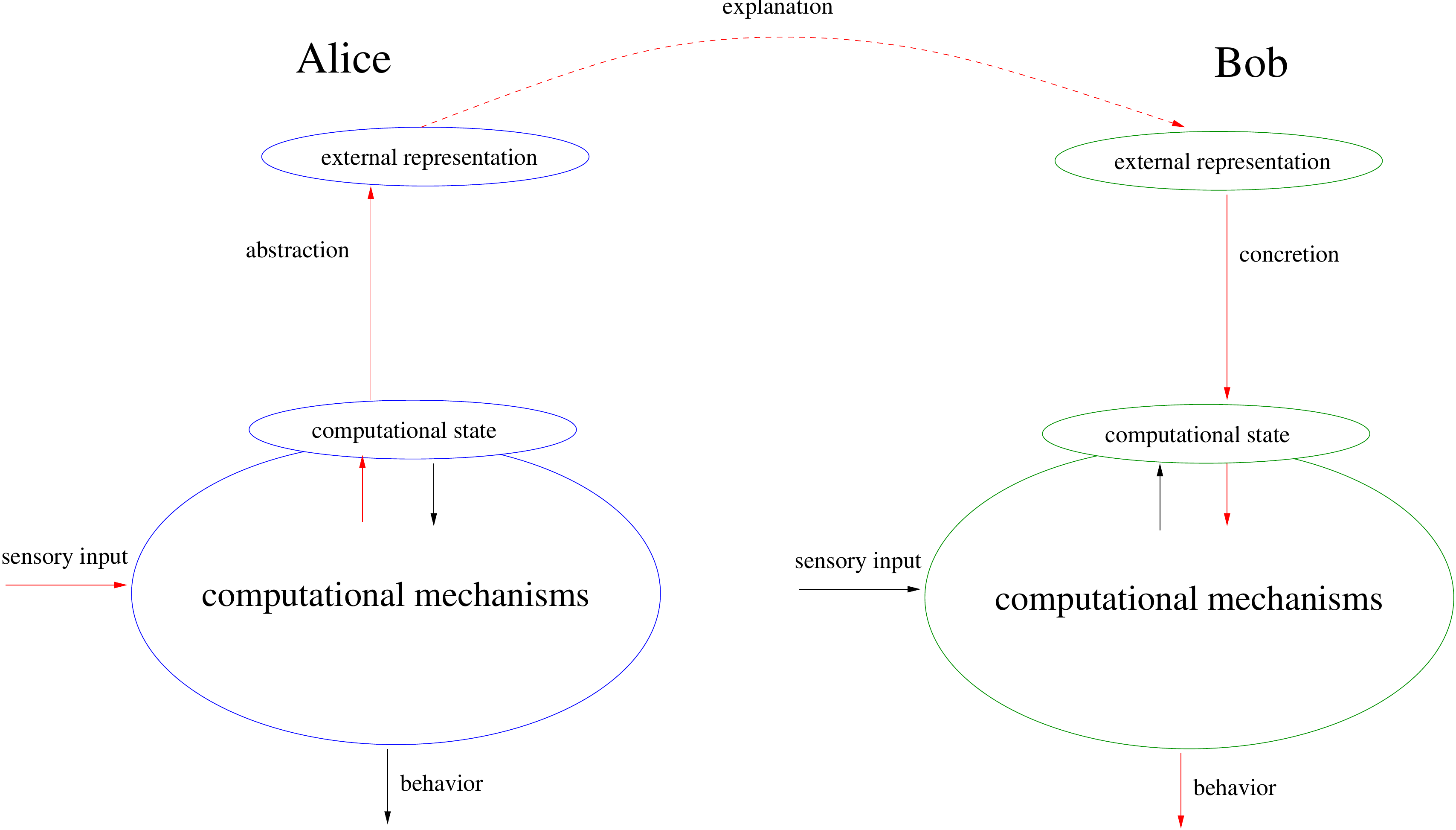}
\caption{\label{modelpic1}Architecture of Mechanism Necessary to Create Shared Intentionality}
\end{figure}

Thus, what is transferred cannot be a representation or description of
the implementation level of the agent's computational state.  Instead,
it must be abstracted into some representation that is common to both
Alice and Bob.  It needs to be abstracted for two reasons: first, it
must be feasible to communicate it (e.g., by demonstration, mime,
or---later---language), so it must be succinct; second, it must be
common to all participants.  In outline, the strategy for shared
intentionality will then be as follows: Alice computes an abstraction
of relevant aspects of her low-level computational state into a form I
call the ``external representation''; this is communicated to Bob (I
consider how this is done below), whose computation inverts the
abstraction that created this representation, an operation referred to
as ``concretion,'' and thereby enriches his computational state so
that it now contains information and ``ideas'' similar to Alice's and
this may lead him to perform usefully cooperative behavior.  Figure
\ref{modelpic1} portrays this design, and the flow of information (red
arrows)\footnote{The external flow labeled ``explanation'' is shown
dotted because this is a virtual, rather than physical, flow of
information; the physical flow is accomplished by Alice generating
behavior that is sensed by Bob.}  from Alice (on the left, in blue) to
Bob (on the right, in green).  Observe that concretion is performed by
Bob, and is therefore able to target the low-level representation used
for his computational state.

\exmemo{Subconscious or unconscious?  The figure should also say planning/execution.}

We have seen that the external representation cannot be couched in
terms of the physical computational or mental state---that is, in
terms of bits and bytes, or neurons, connections, chemistry, and
signals.  Instead, it must employ some more abstract
vocabulary\footnote{Computer scientists would generally use the term
``ontology'' here, but that usage is nonstandard and somewhat
idiosyncratic to the field, so I prefer the more neutral term
``vocabulary.''}  that is common to Alice and Bob.  If the agents were
built on current technology, Alice and Bob's developers would likely
use unsupervised machine learning to provide them with means to
interpret their perceptions and could choose methods that will
``chunk'' their world similarly (e.g.,
\cite{Tenenbaum-etal11:grow-mind}), even though they might be
developed separately.  Anthropomorphically, I will refer to these
chunks as ``concepts.''

The external representation might be no more than a string of
concepts, but it will be more effective if these can be linked
together in a way that indicates sequencing, intent, or causation.
Thus, in addition to sensed objects (i.e., prelinguistic nouns),
chunking should recognize and the vocabulary should include actions
(verbs) and spatial and temporal relations (prepositions).  The
abstraction mechanism then constructs a ``story'' or, as I will say,
an \emph{explanation} in this vocabulary that suggests a state of the
world and actions on it whose concretion matches the salient part of
Alice's mental state (that is, her ``idea'').  Here, it is likely that
Bob is missing the concept ``bridge,'' so Alice might employ the
concepts ``walk,'' ``on top of,'' ``fallen tree,'' ``across,'' and
``ravine'' and her explanation will link these together in sequence,
and perhaps indicate intent: ``in order to'' ``go to'' ``other
side.''\footnote{Those aware of dialog systems based on ``Large
Language Models,'' such as ChatGPT and its cohorts
\cite{OpenAI:InstructGPT22}, may wonder why I do not invoke a reduced
form of this technology here.  The reason is that our hypothesized
agents are prelinguistic: we are attempting to discover where language
comes from.}

The means whereby Alice conveys the explanation to Bob might be
demonstration, mime, or signs and sounds (repurposed from the
communications built-in for preprogrammed interactions) that by
convention are associated with specific concepts.  For example, she
might use a twig to scrape a small ditch in the dirt, then lay the
twig across it and ``walk'' her fingers across, and then point to the
ravine and indicate the selected log.  Bob will watch this mime and
his sensory-processing faculties must recognize that it has symbolic
or conceptual content, extract the concepts and explanation, and then
concretize them so that they are available to his computational state
and mechanisms.

We have now applied the first two stages of Marr's model: we have
postulated the purpose and strategy of the computations performed by
Alice and Bob in achieving shared intentionality (i.e., use of
abstraction/concretion), and of the representations employed and
communicated between them (i.e., explanations over concepts).  Now let
us consider how Alice might perform abstraction and thereby derive
something of the third or algorithmic stage of description.
Some of the mechanisms I propose, and some that I suppose are already
present in our agents, may seem rather arbitrary, but there is a
purpose behind these choices: they are based on those known or
hypothesized to operate in the human brain.

The goal for our algorithm is to construct an explanation---a succinct
external communication---whose concretion by Bob will reproduce parts
of Alice's computational state.  If Alice is to construct such an
explanation, she surely needs to do it using an estimate of Bob's
concretion operation: hence, she must have what philosophers and
psychologists call a ``theory of mind''
\cite{Premack&Woodruff78,iep-theory-of-mind}.\footnote{Theory of mind,
also known as ``mindreading'' \cite{Baron-Cohen99}, should
not be confused with shared intentionality.  Mindreading infers
another agent's beliefs and intentions by observation of its behavior
in its environment (e.g., by simulating its point of view, or by
applying deduction \cite{Carruthers&Smith:ToM96}); shared intentionality involves
transfer of internal ``ideas'' that cannot be directly observed or
inferred: their communication requires deliberate, symbolic actions,
such as Alice's mime.}  For humans, researchers divide ``social
information processing'' into processes that are relatively automatic
and driven by stimuli, versus those that are more deliberative and
controlled.  These distinctions are reflected in the neural structures
that underlie social cognition \cite{Adolphs09}.  We suppose that
processes similar to the automatic ones are part of Alice's basic
computations and lead her to suppose that Bob is similar to her (so it
is worth trying to communicate with him), and the ``more deliberative
and controlled'' processes are part of the mechanism whose structure
we are attempting to deduce.

It is generally understood that any mechanical or living entity that
interacts effectively with some aspect of the world (its
\emph{environment}) must have a model of that environment
\cite{Conant&Ashby70,Francis&Wonham76}.\footnote{Conant and Ashby
explicitly recognized this must apply to the brain, which seems
remarkably prescient for 1970: ``The theorem has the interesting
corollary that the living brain, so far as it is to be successful and
efficient as a regulator for survival, must proceed, in learning, by
the formation of a model (or models) of its environment''
\cite{Conant&Ashby70}.}  For example, the construction and maintenance
of an ``adequately correct'' model of its environment is currently the
central problem in design and assurance of autonomous systems such as
self-driving cars \cite{Jha-etal:Safecomp20}.  Bob is part of Alice's
environment, so we may suppose that she has a model of Bob's state of
knowledge and beliefs\footnote{Beliefs, Desires, and Intentions (BDI)
are a standard way of organizing some aspects of an AI agent
\cite{Rau&Georgeff97}.  Knowledge is understood as true belief.} and
will use this to guide construction of her explanation: she will use a
different explanation if she believes Bob already has the concept of
log-bridges than if he does not.  So now we ask: how does Alice use
her model of Bob's computational state to guide her abstraction?

Let $s$ denote Alice's computational state; her abstraction operation
$\mathit{Alice}_A$ needs to take the relevant part of her state (i.e.,
that concerning her idea for a log-bridge), which we will denote
$\mathit{idea}(s)$, and deliver candidate explanations.  We suppose that the
``more deliberative and controlled'' aspects of Alice's theory of mind
for Bob reside with the new computational mechanism that we are
attempting to construct, so Alice's abstraction operation needs the
potential to offer many explanations, so that the new mechanism can
pick the one, $e$, that will be most effective.  Thus we have
\begin{equation}
\label{edef}
e \in \mathit{Alice}_A(\mathit{idea}(s))
\end{equation}
and we ask how $\mathit{Alice}_A$ is constructed
and how a suitable $e$ is selected.

Now, Alice has her own concretion faculty (for use when she is the
receiver) and I temporarily propose\footnote{I say ``temporarily''
because I will later suggest a more realistic mechanism, but we do not
yet have the context for its introduction.} that this can be
parameterized by models for different ``points of view'' (i.e.,
theories of mind) and thereby simulate (approximately) Bob's
concretion.  For simplicity of exposition, assume that Bob's
concretion operation and Alice's simulation of it are deterministic
functions.  We denote Alice's operation by
$\mathit{Alice}_C(\mathit{Bob}, e)$, where $\mathit{Alice}_C$ is
Alice's concretion function, the argument $\mathit{Bob}$ indicates
this application is parameterized by her model of Bob, and $e$ is an
explanation.  This function delivers a concretized version of $e$ in
the computational form employed by Alice that represents her estimate
of Bob's interpretation of $e$.  We can use the keyword
$\mathit{myself}$ to indicate application of the native
(unparameterized) concretion function, so that Bob's native concretion
function is $\mathit{Bob}_C(\mathit{myself}, e)$, and this delivers
the computational representation of $e$ used by Bob.

What we want is that $\mathit{Bob}_C(\mathit{myself}, e)$ is an
augmentation to Bob's computational state that is similar in effect to
the relevant part of Alice's state $\mathit{idea}(s)$.  We cannot require
$\mathit{Bob}_C(\mathit{myself}, e) \approx \mathit{idea}(s)$ because the left
side uses Bob's computational representation, while the right side
uses Alice's.  But what we can do is require that Alice's simulation
of Bob's concretion delivers a value (which will be in her
representation) that is close to her state $\mathit{idea}(s)$.  That is
\begin{equation}\label{bobc}
\mathit{Alice}_C(\mathit{Bob}, e) \approx \mathit{idea}(s).
\end{equation}
If we use a function $\mathit{error}$ to measure divergence
between the left and right sides of (\ref{bobc}) then we want to
choose an $e$ that minimizes this divergence.  That is, combining
(\ref{edef}) and (\ref{bobc}):
\begin{quote}
Alice's explanation for Bob of her idea $\mathit{idea}(s)$ is $e$ that minimizes
\begin{equation}\label{constraint}
\mathit{error}(\mathit{Alice}_C(\mathit{Bob}, e), \mathit{idea}(s))
\end{equation}
over all $e \in \mathit{Alice}_A(\mathit{idea}(s))$.
\end{quote}

\noindent
This constraint ensures that Alice chooses a good explanation, given
her model of Bob.  We can suppose that pre-existing mechanisms allowed
her to build a model for Bob (see, e.g., \cite{Vasil-etal20}) and, if
necessary, to refine it in a failure-driven ``dialog'' should it prove
inadequate.

\exmemo{Gotta make sure this reads like a feasible design, rather than
anthropomorphic speculation.}

Solving constraint satisfaction problems such as (\ref{constraint})
usually requires some form of optimizing search.  Fortunately, we may
suppose that Alice already has the mechanisms for such search.  As we
noted earlier, any autonomous agent must use information from its
sensors to build and refine a model of its environment.  The untutored
view is that the model is built from the sensors ``bottom up,'' as
when machine learning is used to build the ``detected objects list''
from the cameras in a self-driving car.  However, this approach is
prone to error and instability, not least because it operates
\emph{anti-causally} \cite{Kilbertus-etal18:anti-causal}.  A better
alternative turns things around and uses the model to generate or
\emph{predict} observations and then uses the resulting
\emph{prediction error} to refine the model
\cite{Jha-etal:Safecomp20}.\footnote{The advantage of generative
models is that they reason forwards, or causally, from (models of) the
world to predicted observations; this can take account of sensor
defects and observer behavior (e.g., displacement of sensors due to
movement) and is more straightforward, simpler, and generally more
accurate than the reverse inference.  Most of the recent advances in
AI, such as the Large Language Models employed with ChatGPT (where the
``G'' stands for ``generative'') and similar systems use generative
methods.  Notice that predictions are not necessarily at the sensor
level (e.g., individual pixels) but can target some limited bottom-up
interpretation of these (e.g., detected objects in a self-driving
car).  Also note that bottom-up interpretation may be used to create
an initial model: this is simply the prediction error when there is no
prediction.}  This can be mechanized using techniques known as
Variational Bayes \cite{Fox&Roberts12} in which inference is achieved
via optimization: here, the model is probabilistic (e.g., probability
distributions over attributes of items in the detected objects list),
predictions correspond to Bayesian \emph{priors}, prediction errors
encode observations, and the Variational Bayes algorithm constructs an
update to the model (the Bayesian \emph{posterior}) that approximates
that needed to minimize future prediction
error.\footnote{Conceptually, this is similar to a Kalman Filter,
applied to complex models.}  Similar processes are thought to operate
(across a hierarchy of models) in the brain (we discuss this later)
where they are described as \emph{Predictive Processing} (PP)
\cite{Wiese&Metzinger:VanillaPP2017} and I will use the same term
here.

\exmemo{Following is not right.  Need to identify prediction with
concretion, abstraction with prediction error, and need to explain how
explanation is constructed using theory of mind.   Also, upper model
is like a cache for explanations.}

We suppose that our agents' basic computational mechanisms (in
particular, their perception systems) use PP to build models of the
environment and these are represented in their computational state.
We then use these same mechanisms to construct explanations.
Specifically, we use ``higher-order'' applications of PP to build
an \emph{abstract model} of the world, based on concepts, from the
models represented in the basic computational state.  The abstract
model acts as a ``cache'' of building blocks for explanations (a
richer form of what we previously called the ``external
representation''), so that these do not need to be built from scratch
each time.

\begin{figure}[t]
\includegraphics[width=\textwidth]{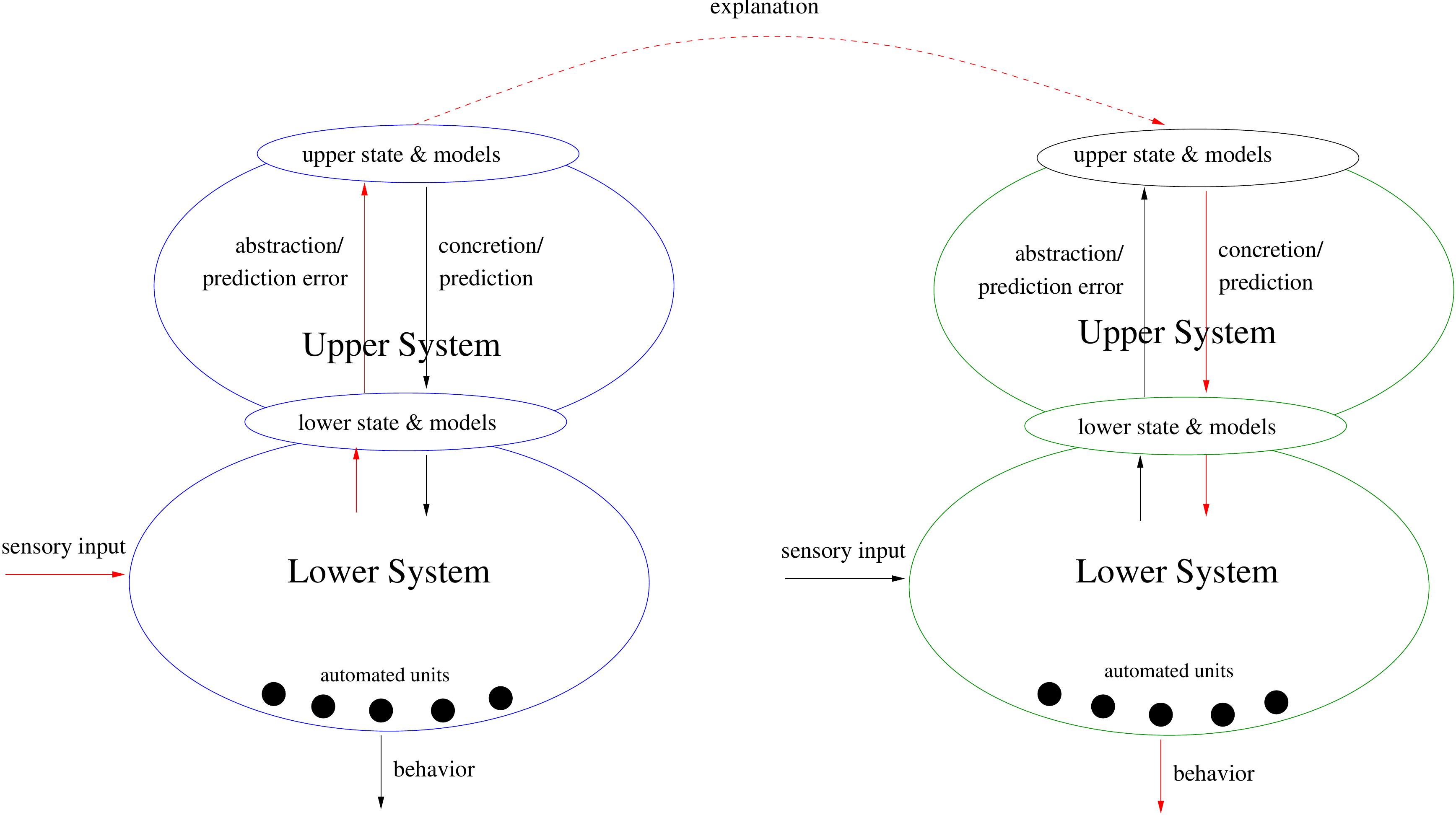}
\caption{\label{modelpic2}Architecture of Implementation to Create Shared Intentionality}
\end{figure}

I will refer to these elaborations of the agent's basic pre-existing
computational mechanisms in Figure \ref{modelpic1}, which now include
construction of world models using PP, and specialized ``units'' that
perform automated calculations on built-in models (such as those for
navigating 3D space), as its ``Lower System.''  And I will refer to the
computational state of this system, which contains representations of
the Lower or ``Concrete Model'' as the ``Lower State.''  Similarly, I
will refer to the new mechanism that uses PP to build concept-based
models of the Lower State as the ``Upper System'' and to its
computational state, which includes the Upper or ``Abstract Model,''
as the ``Upper State.''  Figure \ref{modelpic2} portrays this
implementation of our mechanism for shared intentionality.

Whereas previously Alice repeatedly had to build and optimize new
explanations for each communication to Bob, these can now be
constructed as augmentations to her persistently maintained abstract
model.  The ``deliberative and controlled'' aspect of her theory of
mind for Bob will be part of her abstract model, so that rather than
a separate parameter to her concretion function as supposed earlier,
it simply participates in predictions along with other relevant parts
of that model and, by minimizing prediction error, PP will solve the
constraint previously represented as (\ref{constraint}) and thereby
generate a suitable explanation.

When Bob receives an explanation from Alice, it will first be detected
by his sensors and Lower System.  However, the Lower System will be
unable to interpret its symbolic content and will simply add it to the
Lower State.  His Upper System will not have predicted this addition
to the Lower State and learns about it as a prediction error.  The
Upper System will then extract and interpret the concepts contained
therein, which will enrich its abstract model.  PP then sends a
concretion of this to the Lower System, where it will generate a large
prediction error (since it presumably contains information that is new
to Bob).  Prediction errors can elicit two responses: one is to send
the error to the Upper System, where it may cause revision to the
abstract model maintained there; the other is to change the Lower
State and its concrete model in a way that reduces the error.  This
cannot be done arbitrarily since the Lower State and model must be
consistent with sense data from the environment.  One option is to
adjust the interpretation of sense data (as when we resolve an optical
illusion), and the other is to perform some behavior that will adjust
the environment.  William James was the first to suggest that behavior
is driven by (what we call) the Lower State: ``every mental
representation of a movement awakens to some degree the actual
movement which is its object'' \cite{James:principles}.  Thus, for
example, an explanation may indicate that Bob's right hand should
grasp part of a specific tree.  After concretion, this will be
represented as a configuration of Bob's concrete model that differs
from its current state and this prediction error can be resolved by
Bob actually moving his right hand and grasping the tree in the manner
indicated \cite{Friston:PP09,Friston:free-energy10}.

We have not said much about the Lower System, but in a typical robot
it will contain a collection of automated modeling, planning,
calculation, and execution ``units'' for specific tasks, often
interacting through a ``blackboard architecture''
\cite{Nii86:blackboard}.  The idea here is that elements of the Lower
State are deposited in a common memory or workspace (the
``blackboard''), from where they are removed by those units that
recognize and ``know'' how to interpret them: these units will produce
results that are deposited back in the blackboard for consideration by
other units, and/or they may generate behavior as described above.

This concludes my account of mechanisms for constructing shared
intentionality among computational agents.  I now claim that the
mechanisms assumed and developed here are consistent with those of the
human brain and can explain the emergence of shared intentionality in
humans, as described in the following section.

\section{Human Interpretation}

The mechanisms I have proposed to endow the robots Alice and Bob with
shared intentionality are plausible but probably not those that a
robot designer would use to deliver such capability today: there are
more powerful technologies that can create shared intentionality in
robots by direct communication of models, goals, and plans in some
pre-arranged shared format (i.e., something closer to a language).
Nonetheless, the proposed mechanisms are perfectly feasible and I
chose them because they are prelinguistic and based on capabilities
known or generally considered to be present in the human brain:
specifically, predictive processing, a dual-process architecture with
powerful and autonomous low-level automation, and some form of global
workspace.  Thus, I propose that aspects of shared intentionality (or,
more generally, ``collective'' intentionality
\cite{Searle:collective-intentions90}) in humans are created by the
same mechanisms as those described for Alice and Bob.  Of course, this
assumes that evolution provided some of our ancestors (I will call
them ``proto-humans'') with capabilities similar to those of Alice and
Bob, but not yet with anything more powerful for the direct
construction of shared intentionality, like language.  I think this is
plausible, because (temporally adopting teleological usage) there is
no reason for language to have evolved prior to the construction of
shared intentionality.

I will say more about evolutionary plausibility in Section
\ref{evolution} when I have described additional capabilities that I
believe are associated with shared intentionality.  I will argue that
these related capabilities build on the mechanisms for shared
intentionality and therefore it was the first to emerge: hence, I call
this the Shared Intentionality First Theory (SIFT) for emergence of
these capabilities.

The mechanisms developed for shared intentionality in Alice and Bob
make extensive use of predictive processing (PP).  This is a popular
theory of brain operation \cite{Wiese&Metzinger:VanillaPP2017}, where
it is also known as ``predictive coding'' \cite{Clark13}, ``predictive
error minimization'' \cite{Hohwy:book13}, and (using terminology
derived from statistical physics) the ``free energy principle''
\cite{Friston:PP09}.\footnote{Some authors (e.g.,
\cite{Doerig-etal21,Hohwy&Seth20}) treat PP as a theory of
consciousness in itself, whereas I regard it as a basic mechanism of
perception that is probably present in all animals with a nervous system and
brain.}  Its use in computer science is to a large extent inspired by
these biological precursors.

Recognition that sense interpretation must work ``top down'' rather
than ``bottom up,'' as conventionally assumed, was first documented by
Helmholtz in the 1860s \cite{Helmholtz1867} and developed in more
detail by Gregory \cite{Gregory80} (who, in the 1980s, explicitly
related perception to hypothesis testing in science), and by Rao and
Ballard \cite{Rao&Ballard99} in the 1990s.  PP posits that the brain
builds probabilistic models of its environment and uses these to
predict its sensory input.  The predictions are compared to sensed
reality and the differences are used to refine the models via (an
approximation to) Bayesian variational inference in a way that
minimizes prediction error.  This minimization can be achieved by
refining either the upper or the lower model, or by changing the
environment; the latter may be achieved by using our body to perform
actions \cite{Friston:free-energy10}, as described earlier.

PP has Bayesian priors flowing from models down to sense organs as
predictions and observations flowing back up as prediction errors, and
this explains the otherwise puzzling fact \cite{Graziano13}
that there are many more neural pathways going from upper to lower
levels of the brain than vice versa (predictions require more
bandwidth than errors).  PP in humans differs from that described for
Alice and Bob in that the human perception system maintains many
levels of intermediate models, each contributing a small step to the
overall interpretation (e.g., edge and motion detection for vision
\cite{Rao&Ballard99}, different time scales for speech
\cite{Caucheteux23}), whereas Alice and Bob's perception systems have
just the single lower level.  However, their sensor interpretation at
that lower level will be implemented by deep neural nets whose layers
will each build representations that could be regarded as intermediate
models---but note that these representations are refined only during
training and are fixed thereafter, whereas intermediate human models
are under constant revision.

Beyond the multiple models of its perception system, the human brain
is postulated to have a ``dual-process'' macro-scale architecture
comprising ``System 1'' (fast, automatic, prone to error) and ``System
2'' (slower, requires direction and effort, can perform reasoning)
\cite{Frankish10,Evans&Stanovich13}, popularized as ``thinking, fast
and slow'' \cite{Kahneman11}.  Note that this is a logical
architecture; it need not be realized as physically separate parts of
the brain.  The perception system's lower-level models are found in
System 1 and higher-level ones in System 2.  Alice and Bob likewise
have Lower and Upper Systems with corresponding concrete and abstract
models.  In the following, I will suggest how rationality emerges
within these dual-process architectures,

\exmemo{Reprise lower units and metaphors: these are explained in the
next section.}

\subsection{Local Impact of Shared Intentionality Mechanism: Rationality}
\label{rationality}

My earlier account of predictive processing was incomplete in that I
described prediction errors leading to incremental model refinement.
In fact, this occurs only with ``small'' prediction errors; ``large''
prediction errors indicate a ``surprise,'' meaning the world has not
evolved as expected, or the performance of our sensors has changed or
failed (e.g., dazzled by the sun).  Surprise causes a more drastic
reappraisal of models and plans; the extent and effectiveness of the
reappraisal obviously depends on the ``reasoning'' capability
available.  The mechanisms I have proposed for shared intentionality
have endowed our proto-humans with an Upper System that maintains an
abstract Upper Model over which they are able to construct and
manipulate explanations.  These capabilities provide our proto-humans
with a sophisticated response to surprise and also form a
foundation for independent reasoning.

\begin{figure}[t]
\begin{center}
\includegraphics[width=2.8in]{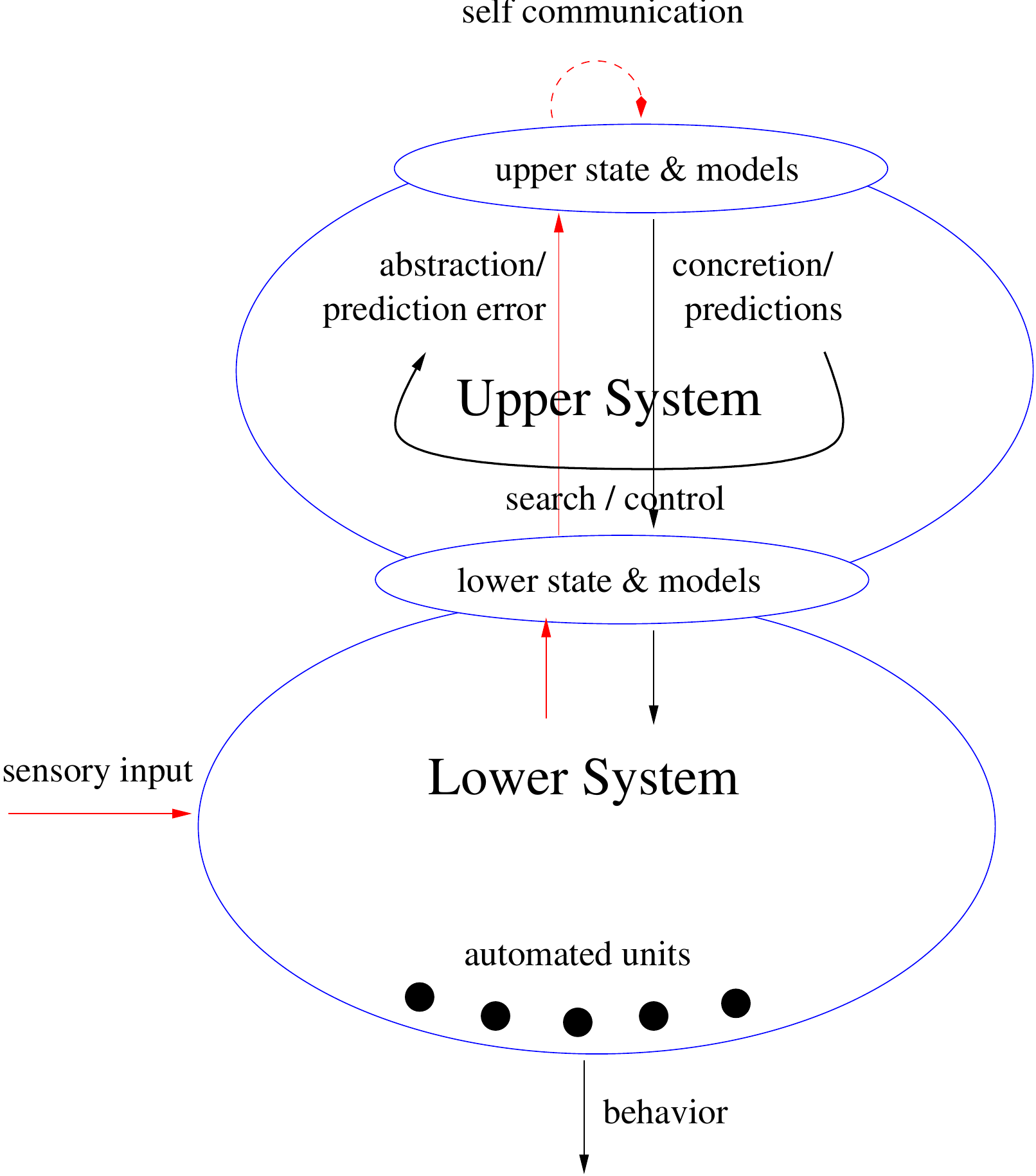}
\end{center}
\caption{\label{solo}The ``Local Loop'' for Rationality}
\end{figure}

In particular, just as Alice can communicate with Bob to create shared
intentionality, so she can also communicate or ``talk'' (wordlessly)
to herself.  Her faculties for abstraction, explanation, and
concretion can be employed in a local loop, interacting with the
built-in automation provided by specialized units in her Lower System.
This is portrayed in Figure \ref{solo}, although in reality the
self-communication is internal.

The power of this architecture is that the Upper System has the
ability to construct and deconstruct explanations: that is, it can
manipulate relations among concepts.\footnote{We must be careful to
postulate only limited ability here; otherwise we are invoking
something close to language.  What I have in mind is the
``protolanguage'' of Bickerton \cite{Bickerton:species90} (i.e.,
language lacking grammar) or the elementary
``Isolating-Monocategorial-Associational'' (IMA) grammar of Gil
\cite{Gil:IMA06}.  Young children, and pidgin speakers, are able to
accomplish quite a lot with just such primitive combinations of words.}
Freed from the need to construct a communication for Bob, Alice may be
able to exploit and control the search for explanations in new ways.
For example, by manipulating her model of his concretion, she can
perform counterfactual and ``what if'' reasoning.

I propose that this local loop is the basis for human rationality, by
which I mean the construction of plans and actions that may be
expected to achieve their objectives despite an uncertain world.
In operation, the Upper System might construct a conceptual goal; its
local capabilities for manipulating concepts may rearrange or
decompose some elements of this and concretion can send some of them
to specialized units in the Lower System, where they will be
transformed in other ways and then abstracted back to the Upper System
where they might be further manipulated to yield an explanation that
solves the original goal (which can be checked by concretizing both
goal and explanation and applying them to models of the world
maintained by the Lower System).  Alternatively, the goal might
originate in the Lower System, but the loop will operate in a similar
way following an initial abstraction to the Upper System.

The local loop is not just a problem solver: it can produce action.
When Alice receives a communication from Bob, its concretion augments
her Lower State and thereby influences her future behavior (recall
earlier discussion that divergence between the sensed environment and
Lower Model may be resolved by actions that change the environment).
And this will be true of Alice's rational deliberations, too: the
local loop can construct a solution to an upper- or lower-level goal
and its concretion will set the Lower System on course to deliver
suitable behavior.

As noted, concretion allows the Upper System to recruit specialized
Lower System units as subroutines.  Modern computerized reasoning
systems do something similar, using highly efficient units such as
``SMT Solvers'' as subroutines \cite{DeMoura&Bjorner11}.  The native
problem is first transformed into a ``Satisfiability Modulo Theories''
(SMT) problem, solved by the SMT Solver, and then the result is
translated back to the terms of the original problem.  The Upper
System of proto-humans can likewise use a Lower System unit that
models vertical space as a subroutine in reasoning about social
hierarchy.\footnote{The Lower System or subconscious in humans is
thought to have numerous specialized faculties, although their
identity is open to debate.  Evolutionary psychology might argue for
mental ``modules'' such as those for social rules, mate selection, and
so on \cite{Pinker:how-mind03}, while embodied cognition might argue
that specialized calculation derives from our interaction with the
physical world, and concerns reasoning about distance, weight, time,
etc. \cite{Jostmann-etal09}.  For my purposes, it is sufficient to
acknowledge the existence of specialized cognitive units, without
worrying about their precise functions.}  When expressed
linguistically, the transformations become metaphors: ``he is my
superior and she is my peer, but I am above the others.''  Lakoff and
Johnson \cite{Lakoff&Johnson08} observe that this application of
metaphor pervades our thinking: it is a primary mechanism of thought,
not just a feature of language, and now we can see why (and that it is
prelinguistic).\footnote{Lakoff \cite{Lakoff-mapping14} describes
mental mechanisms for metaphor that are more complex than mine, but
they are not incompatible.}

I think this proposed architecture also illuminates the ``fast and
slow'' dual-process model \cite{Kahneman11} and suggests why
deliberative thinking is slow and easily fatigued: the ``fast''
mechanisms use the specialized units built-in to the Lower System that
operate ``in parallel,'' whereas the ``slow'' ones employ rather
costly local loops, optimizing searches, and transformations that
operate as a ``single thread.''  As Kahneman illustrates, and perhaps
due to its costs, this architecture for rational deliberation does not
guarantee good results: the Lower System automation is often ``rough
and ready'' and the Upper System might use it poorly via an
inappropriate metaphor; furthermore, its consideration of alternative
and unfavorable scenarios may be optimistic and cursory.  Thus, it
requires effort to derive full benefit from this capability.  That
effort is composed of attention and control, which are the foundations
for intentional consciousness, as we will now consider.

\subsection{Intentional Consciousness}

Consciousness is a notoriously difficult topic, with many facets.
However, it is generally agreed that two of these are primary:
\emph{intentional} consciousness,\footnote{Also called \emph{access}
or \emph{cognitive} consciousness.} and \emph{phenomenal}
consciousness.  The first concerns the ability to direct attention and
to think \emph{about} something and to know that you are doing so.
The second concerns ``what it's like'' to have subjective experiences
such as the smell of a rose or a feeling of pain, experiences referred
to generically as \emph{qualia}.

We started this investigation by considering mechanisms that can
construct shared intentionality, so it will not be surprising if these
mechanisms also deliver individual intentionality and, thereby, this
facet of consciousness.  Let us consider how this might come about.

When a proto-human Alice constructs an explanation to communicate to
Bob, or engages in rational deliberation using a local loop, she
surely needs to focus her resources on these tasks and on the
explanations generated and received.  The lower or subconscious system
state records and represents a vast amount of information from all our
sense organs, models of the world, memories, beliefs, desires,
intentions, and all the ongoing automated processing of these.  The
abstraction mechanism must be highly selective about what of this it
chooses to represent in its upper model and to use in constructing its
(one) current explanation: this selectivity and focus is what we mean
by \emph{attention} and my proposal is that intentional consciousness
corresponds to \emph{awareness} of attention and of the things
attended to in the Upper System.  In fact, I suggest that intentional
consciousness goes beyond awareness of attention and also has elements
of executive control, so that we can focus our resources on particular
deliberations (unfocused deliberations will interfere with each other
and make little progress).  This focused deliberation seems different
than the automated mental processes of the Lower System, which
operates its units ``in parallel'' (i.e., simultaneously), weighted
and directed according to the exigencies of the moment.  For our Upper
System deliberations, we are largely able (and required) to maintain a
directed focus that seems, despite occasionally ``wandering,'' to
operate as a single ``sequential'' thread that is ``about'' some topic
or goal.

This single-threaded purposeful focus of attention and control, and
awareness of it, seem sufficiently different than other mental
processes that it should not be surprising that it corresponds to a
unique mental experience which, I propose, is intentional
consciousness.\footnote{This contrasts with Dennett's ``multiple
drafts'' theory \cite{Dennett:explained93}: I accept there may be many
fleeting drafts or models at the subconscious level, but consciousness
resides with the one explanation that is currently a candidate for
communication or the subject of local deliberation.}  However, although this
explains what the computational processes underlying
intentional consciousness are for and how they are constructed, I
cannot explain how they produce an apparently nonphysical
experience.\footnote{Observe, too, that in constructing an explanation
for Bob, Alice surely needs to recognize herself as distinct from him.
Hence, I speculate, but once again cannot explain, that these
processes also produce a ``self-model''\!\cite{Metzinger04} and the
illusive experience of personal identity \cite[Book\,I.iv, section
6]{Hume:treatise}.}  This is a problem for all materialist theories of
consciousness and, unlike Graziano \emph{et al.}
\cite{Graziano-etal20} and others, such as the ``illusionist'' school
\cite[whole issue]{Frankish16}, I do not think the problem is solved
by claiming that these processes cause the brain merely to
\emph{report} or to register \emph{belief} in a nonphysical
experience.  Thus, I recognize that a leap of faith is required to
accept this interpretation of intentional consciousness, but I argue
that it accords with observation, and with some other theories of
consciousness.

First, I need to expand our previous considerations of shared
intentionality and rationality, where I implicitly supposed that the
mechanisms of the Upper System (i.e., abstraction, concretion, and
manipulation of explanations) are invoked periodically to accomplish
specific acts of communication or rational deliberation.  Now I
suggest that these mechanisms do not otherwise sit idle, but operate
continuously and autonomously and we attend to them deliberately, and
are conscious of them, only when focus is necessary.  As life
proceeds, the subconscious Lower System generates behavior, sometimes
at the behest of the Upper System but often on its own.  This is
constantly monitored by the Upper System, which maintains an abstract
model and generates explanations whose concretion closely tracks the
Lower System model and state.  Although our behavior is mostly
generated by the Lower System, we are conscious only of the
abstractions and explanations reconstructed by the Upper System; thus,
it ``feels'' as if consciousness causes behavior.

This explains some otherwise puzzling facts.  For example, experiments
such as Libet's \cite{Libet85}, and studies with ``split-brain''
patients \cite{Gazzaniga:who12}, reveal that, contrary to our
intuitions, the conscious mind is less an initiator of actions and
more a reporter and interpreter of actions and decisions initiated in
the subconscious.  But we can now see that the primary purpose of the
mental faculty that supports consciousness is precisely the reporting
and interpretation needed to construct shared intentionality;
rationality and consciousness ride on the mechanisms of shared
intentionality and share its character.

Next, the experience of intentional consciousness is derived from the
Upper System, whose foundational purpose is to construct communicable
prelinguistic abstractions that can deliver shared intentionality.
This explains why much of the experience of intentional consciousness
is of an inner dialog (rather than, say, a stream of images).
Initially the dialog would be wordless because it is difficult to
imagine how speech could have evolved prior to shared intentionality,
but language, and ultimately spoken language, could surely have
developed quite rapidly once shared intentionality became available:
the ``gist'' that is at the heart of language understanding and memory
\cite{Brainerd:gist90} could have evolved from the wordless
concept-based explanations of shared intentionality, which might also
be a basis for ``mentalese,'' the ``language of thought''
\cite{sep-lang-of-thought}.

To consider further observations that can be explained by this theory
of intentional consciousness, we need to see how phenomenal
consciousness fits in.

\subsection{Phenomenal Consciousness}

When Alice explains her idea for a log-bridge to Bob, she might finish
by pointing to a particular log as the one to be used.  This is a
remarkable thing: she is making an external reference to a
subjective inner experience, namely her visual field.

Once we have the ability to construct shared intentionality, we need
the additional ability to communicate our perceptions of things in the
world and things about ourselves.  We experience these through our
senses and so to reference them in communications it is necessary for
subconscious information derived from our senses to be abstracted into
the Upper System.

I submit that this is phenomenal consciousness: in order to
communicate the things we sense and feel, selected abstractions and
explanations about them must be present in the Upper System and its
model---and we will be conscious of them just as we are conscious of
other content in our abstractions and explanations.  It is one thing
for our visual system to allow us to sense a log that we may choose to
sit down on---all this can be done subconsciously and ``in the dark,''
as it is by people with ``blindsight''
\cite{Humphrey:blindsight06,Derrien-etal22}---and quite a
different thing for our phenomenal consciousness to present us with
the experience of the visual field, so that we can indicate ``the log
on the left.''  Notice that this indication is symbolic, it is not a
direct response to the sensation concerned, as when we sit down on a
log or recoil from pain.  And notice, too, that consciousness of the
relevant sense arises only because we may need to communicate it to
others.

We have senses beyond the classic five, possibly as many as 50
\cite{Ward:sensational23}; it is notable that we are conscious of some
of these---and to different degrees---and others not at all.  For
example, the sense of balance (kinesthesia) is important and we are
conscious of being in or out of balance, but we are not conscious of
the sense organs (e.g., semicircular canals) that support kinesthesia
(except their \emph{mal\,}function, as when we feel dizzy).  And for
another: I have a colleague whose proprioception is failing and he
finds it difficult to describe the symptoms because this sense is
largely unconscious and consequently we have no vocabulary for ``what
it's like.''

It seems that the senses of which we are phenomenally conscious are
just those that it can be useful to communicate explicitly to others,
notably including the classic five.  We are not phenomenally
conscious of the senses supporting kinesthesia because we do not need
to communicate their detailed content (e.g., ``the current rate of yaw
is 5 degrees per second'').  So there is little ``that it's like'' to
experience the senses supporting kinesthesia---and nothing at all for
proprioception.  Hence, I think it is quite possible that there is
nothing ``that it's like'' to be a bat \cite{Nagel74} because,
assuming bats do not create shared intentionality, they lack an Upper
System and all their senses and actions operate nonconsciously, ``in
the dark.''

In addition to our senses, we need to communicate certain subjective
experiences such as hunger, thirst, and pain (as a sensor for injury
or sickness), together with moods and emotions, so these must be
abstracted to the Upper System as well.  It does not particularly
matter how the color of a red rose is represented in the Upper System,
nor the anguish of jealousy (i.e., ``what it's like'' to experience
these), as long as they can be distinguished.

This proposal partially solves the ``hard problem'' of consciousness
\cite{Chalmers95:hard-problem}: why are some phenomenal states
conscious?  They are conscious because we need to communicate them to
others and so their abstractions must partake in explanations and be
present in the Upper Model, attended aspects of which are conscious.
I say ``partially solves'' because it explains what the process
underlying phenomenal consciousness is for, how it works, and why it produces a
nonphysical experience---but it does so by establishing a relation to
intentional consciousness and, although I have also explained what
that aspect of consciousness is for and how it is constructed, I
accept that I cannot explain why or how it produces a subjective
experience.

Furthermore, this proposal does not explain why our subjective
experience of, say, red is what it is and not something else.  I am
unapologetic about this: my opinion is that it is an unanswerable
question, and an unimportant one: as stated above, red has to have
some representation in the Upper System so that we can reference it
(``the mailbox is the red thing''), but the form of that
representation and ``what it's like'' do not matter, provided they are
distinct from those of other perceptions.  Neither does it explain
``what it's like'' for different people to experience similar qualia:
is your perception of red the same as mine?  There seems to be no
requirement for this to be so and the existence of synesthesia
\cite{Grossenbacher&Lovelace01} or the phenomenon of ``The Dress''
\cite{the-dress-wiki} indicate that people can indeed
experience the same qualia differently.\footnote{The Berlin-Kay theory
\cite{Berlin-Kay91}, whereby the selection of basic color terms in a
culture is predicted by the number of such terms, suggests to me that
mostly we do experience qualia similarly.}

Many will find this disappointing: phenomenal consciousness is the
experience most central to our lives.  But the fact that its
representation is arbitrary does not diminish its significance.  What
matters is that we \emph{care} about the form that it does take, an
attribute sometimes referred to as \emph{sentience} \cite{Duncan:sentience06}.

A less discussed variant on this question asks why our phenomenal
consciousness occurs at the representational level that it does
\cite{Jackendoff87}.  Vision, for example, employs dozens of
specialized units that build representations ranging from a ``primary
sketch'' at the lower levels (which are unconscious), through the
shaded, 2.5D, photograph-like ``intermediate level'' (where my
phenomenal consciousness currently sees a black parallelogram on top
of a partially occluded larger brown parallelogram) to a top-level
conceptual model (where I recognize my keyboard on top of a cluttered
wooden desk).  Jackendoff \cite{Jackendoff87} observed that the
intermediate level (as identified by the theories of sensory
interpretation current at that time \cite{Marr82}) seem to be where
phenomenal consciousness is located.  Prinz \cite{Prinz17} argues that
this observation remains true under modern theories, and the question
is why is it this level, and not some higher or lower one?

Marchi and Hohwy attempt to answer this using a PP model of brain
operation \cite{Marchi&Hohwy22}.  They argue that it depends ``on the
spatiotemporal resolution of the typical actions that an organism can
normally perform.''  For humans, this makes the intermediate level
appropriate, but this is ``not an essential feature of consciousness;
in organisms with different action dispositions the privileged level
or levels may differ as well.''

I speculate that SIFT provides a simpler explanation: the
representational level of consciousness is whatever was the top level
prior to the evolution of the Upper System.  In Figures
\ref{modelpic2} and \ref{solo}, the Upper System abstracts from and
concretizes to what is labeled as the ``Lower State \& Models.''  My
speculation is that these models correspond to the (collection of)
top-level representations prior to the evolution of the Upper System.
These representations are not themselves conscious; phenomenal
consciousness resides in the Upper System's higher-level
representations of (or references to) these.  Viewed relative to the
high-level conceptual models built by the Upper System, the target of
phenomenal consciousness is indeed an intermediate level, but that is
a consequence of the subsequent evolution of the levels above it.

\subsection{Biological \& Evolutionary Plausibility and Evidence}
\label{evolution}

I have advanced a proposal for shared intentionality, rationality, and
consciousness that I call the ``Shared Intentionality First Theory''
or SIFT\@.  The proposal is that these mental attributes form a
``package'' but that shared intentionality provides the framework on
which the others are constructed.  The proposal is based on
computational constructions for hypothesized humanoid agents or
robots.  However, the constructions assume an underlying computational
architecture with capabilities selected from those known or believed
to exist in humans.  These include predictive processing and a
dual-process architecture with multiple specialized ``units'' for
automated lower-level calculation, likely organized around a global
workspace.

This proposal is entirely abstract and computational: I assume that
the purpose and function of the human brain is to perform
calculations, notably those concerned with construction and
interpretation of models of its environment.  I join others
\cite{Brown:tale14} in maintaining that progress in understanding
emergent properties such as consciousness requires abstraction and
theories in addition to experimental neuroscience and, eventually,
development of bridge laws between these points of view
\cite{Kopetz-etal:emergence16}.  As yet, I cannot identify biological
mechanisms or structures or ``neural correlates'' that correspond or
bridge to my overall proposal,\footnote{Recent papers that look at
neural correlates of shared intentionality (e.g.,
\cite{Fishburn-etal18}) focus on the synchronization or ``coupling'' 
of minds rather than active communication between them.}  but there are papers that
do so for its constituent parts (e.g., \cite{Hohwy&Seth20},
\cite[section on Neuroscientific Evidence]{Evans&Stanovich13}).  Thus,
I posit that my proposal is biologically plausible.  I further posit
that it is plausible that the overall package of capabilities evolved
from these constituents.

Humans have shared intentionality whereas other primates do not
\cite{Call09}, so this capability emerged sometime in our recent
evolutionary history and we can ask whether it emerged before, after,
or with related capabilities such as language, rationality, and
consciousness.  The basis of SIFT is that the mechanisms of shared
intentionality evolved first, or provided the survival and
reproductive advantage that caused natural selection to favor the
package.  This differs from other theories of rationality and
consciousness, which go wrong, in my view, at their first step: they
assume, as is natural, that since consciousness is subjective and
personal, it must do something for the individual.  This is not to
deny that consciousness and rational deliberation have benefit for the
individual, only that their evolutionary origin lies in shared
intentionality leading to teamwork that delivers advantage to the
group \cite{Angus&Newton15}.  I am aware that group selection is a
contested notion; however, once this package of capabilities exists,
evolutionary selection can operate on its components for individual
advantage.

A key requirement of SIFT is evolution of an Upper System that
abstracts the Lower State in terms of concepts and can manipulate
those concepts to form explanations.  Some will find circularity in
this invocation of concepts: to communicate we need a shared abstract
vocabulary, which I associate with concepts, but how did these arise
without communication?  

As we have noted before, it is generally understood that the function
of the brain is to guide an animal's interaction with its environment
\cite{Clark13} and to do this it must build models of that environment
\cite{Conant&Ashby70}.  The models cannot be in terms of raw sense
data; even simple animals must perform some categorization on that
data: they must surely distinguish rocks from plants, and plants from
animals, and their own species from prey and predators.  Zentall
\emph{et al.}\ \cite{Zentall08} provide a survey of concept learning
in animals and conclude that similar underlying processes apply to
humans.  Carey \cite{Carey09:origin-concepts} understands human
concept formation to occur on two levels; core concepts (which are all
we require here) are acquired rapidly and early in childhood by the
processes mentioned above, and language is required only for
higher-level concepts, such as ``The United Nations.''  Thus, human
core concept formation could have achieved detailed categorization of
the natural and social world prior to development of the mechanisms of
communication developed here.  However, for communication, and the
construction of shared intentionality, it is not enough to have
concepts: they must be held in common; when Alice points to a tree,
Bob must think ``tree'' not ``leaves.''  Fortunately, there do seem to
be prelinguistic mechanisms that ensure core concepts are shared
among members of a local community \cite{Stolk-etal16,Vasil-etal20}.

Thus, I maintain it is feasible that the Upper System evolved to
perform the functions of shared intentionality---but it is also
possible that these functions were adapted from some prior
dual-process architecture that evolved for other reasons (e.g., to
manage ``surprise'').  Either arrangement suits my purpose (though the
former suggests that the Upper System is unique to humans while the
latter does not).  Cognitive structures such as these leave no
physical evidence in the fossil record so we must look to
archaeological and anthropological evidence of behavior to see if
shared intentionality did emerge first among the package that includes
rationality and consciousness.

It seems there is evidence for collective hunting by ancestral humans
going back millions of years \cite{Guardian12:hunting}, but it is not
clear whether this indicates shared intentionality or merely a
built-in program for group behavior like that of wolves.  More
definite signs of shared intentionality, such as living in large
groups,\footnote{In small groups, everyone knows everyone else and some
form of group behavior can develop based on individual and collective
relationships; in large groups, we need rules, and these need shared
intentionality.} are seen in modern and possibly archaic humans
dating back a few hundred thousand years
\cite{Tomasello14:thinking-history,OMadagain&Tomasello22}.

For signs of rationality and consciousness, I believe we have to look
much later, to the ``explosion'' of creativity (cave paintings, hand
prints etc.)  seen in the human record about 40,000 years
ago.\footnote{These records were first found in Europe and dated to
around 20,000 years ago; more recent investigations have found
precursors in Indonesia dated to 40,000 years ago and in Africa to as
long as 100,000 years ago.  All of these are later than 
emergence of shared intentionality.}  The archaeologist Steven Mithen
attributes this new behavior to integration of formerly separate
cognitive domains \cite{Mithen96} and this would be consistent with
emergence of the ``local-loop'' mechanism for rationality described in
Section \ref{rationality} that is able to exploit multiple automated
Lower System ``units'' (via metaphors).  It is contested whether
Neanderthals, who became extinct soon after this date, engaged in
symbolic thought \cite{Sykes:kindred20}, and consideration of how
their behavior and mental attributes differed from modern humans would
be interesting from a SIFT perspective.  There is clearly opportunity
for more inquiry and evaluation of evidence here, but it does seem
that the human evolutionary and archaeological record may support, and
certainly does not contradict, the theory of Shared Intentionality
First.

Another way to seek evolutionary evidence would be to look for
precursors to shared intentionality, rationality, and consciousness
among living species \cite{Durdevic&Call:origins-mind22}.  Some
attribute shared intentionality to social animals such as wolves, and
even bees, whereas others claim their collective behavior emerges from
simple rules, preprogrammed by evolution \cite{Duguid&Melis20}.  I
join with the latter and believe that one individual of these species
cannot communicate a new idea or plan to another, save by
imitation.\footnote{Domesticated dogs are an interesting case because
it is possible they can participate with humans in ``asymmetric''
shared intentionality: that is, humans may be able to communicate a
goal or plan to dogs, but not vice versa.  There is some evidence that
dogs understand human intentionality \cite{Schunemann:dogs21}, so it
is possible that a symbolic utterance such as ``fetch'' (the ball) is
processed this way, but it could also be a conditioned reflex.
Furthermore, dogs have been bred selectively by humans for thousands
of generations, so it is possible that their mental faculties have
been selected along with their appearance and behavior, and do not
represent the capabilities of dogs in the wild.}
Other animals, even primates, show few signs of shared intentionality
\cite{Call09,Duguid&Melis20,Graham-etal20} but some are popularly
believed to be conscious
\cite{Consciousness:Robson21,Allen&Trestman17}, which would contradict
SIFT\@.  However, we know from Libet's experiment \cite{Libet85} and
its successors that humans attribute behaviors to intentional
consciousness that actually originate in the subconscious.  I suspect
that many of our behaviors are like this, and that when we see similar
behaviors in animals, we attribute them to consciousness because we
falsely believe that is how it is with us.  Thus, although there are
opportunities here for tests and possible refutations of SIFT, my
belief is that consciousness evolved sufficiently recently that it is
not to be found among our ancestor and sister
species.\footnote{Blindsight in apes does provide a possible
contradiction to SIFT, since it suggests that normally sighted individuals are
conscious of their visual field \cite{Humphrey:blindsight06}.}

Hence, I suggest it may be more productive to look for SIFT-like
developments through convergent evolution among hypersocial species
such as elephants, toothed whales, and corvids.  Certainly, sentience
and possibly consciousness are sometimes attributed to these
\cite{Consciousness:Robson21} and it might be enlightening to
investigate their capacity for shared intentionality.  Octopuses are
another interesting and challenging case, as they are widely thought
to be intelligent and possibly sentient, yet they are not social
\cite{Godfrey-Smith16}.  However, each arm of an octopus is capable of
autonomous behavior and has a concentration of neurons somewhat like a
brain; together, these contain twice the neurons of the main brain
(put another way, each ``arm brain'' is a quarter the size of the main
brain) \cite{Carls-Diamante22}.  The neural pathways from the main
brain to those in the arms are too small for high-bandwidth
integration, so it is possible that a single octopus instead creates
shared intentionality (possibly of the asymmetric variety discussed
previously with regard to humans and dogs) among the ``community'' of
its nine separate ``brains'' \footnote{Jennifer Mather disputes this
and likens the arms to ``subroutines'' of the main brain \cite[figure
1]{Mather&Dickel17}, but I think there remains the question of how the
``remote procedure call'' is communicated.} and that SIFT then
delivers more advanced capabilities, possibly including consciousness
(which could be very different to that of humans---lacking unity
\cite{Mather:unity21,Carls-Diamante:unity17}, for example).  There may
be opportunities for research here.

It is shared intentionality and intentional consciousness that create
a r\^{o}le for phenomenal consciousness; thus SIFT predicts that
phenomenal consciousness evolved with or later than shared
intentionality, and that would imply it is unique to humans (plus
possibly those animal candidates for shared intentionality mentioned
above).  This is contradicted by those who believe it is part of the
basic mechanism of advanced perception and arose 500 million years ago
(in the Cambrian explosion) and is possessed by all vertebrates
\cite{Feinberg&Mallatt16}.\exfootnote{When we observe animals behave in
ways similar to our ourselves (e.g., the response to pain), it is
natural to assume they feel (i.e., have phenomenal consciousness of)
the experience as we do.  But this is unwarranted; even bacteria
exhibit aversive behavior in the presence of a noxious stimulus, and
it is perfectly plausible that animals' response to injury or harm is
performed without conscious experience of pain \cite{}; others
disagree \cite{}.}  Evaluating these competing theories is complicated
by lack of any accepted means for assessing phenomenal consciousness
in animals.  Of course, it may be that precursors to the components of
SIFT evolved at different times and in different orders to emergence
of the finished package, and there may be opportunities for
falsifiable experiments here.

Independently of evolution, we could look for direct evidence in
modern humans for some of the mechanisms I have hypothesized.  For
example, infants can provide an opportunity to evaluate shared
intentionality versus consciousness in prelinguistic humans
\cite{Moll-etal21} and the impact of shared intentionality on
cognitive development \cite{Tomasello&Carpenter07}.  These
investigations must be driven by precise hypotheses and, since my
proposal is new, much of it remains work for the future.  However, at
least one relevant capability has been observed in adult humans: this
is the ``interpreter module'' identified by Gazzaniga \cite[Chapter
3]{Gazzaniga:who12}.  This module selectively attends to what is going
on elsewhere in the brain and retrospectively constructs explanations
for the beliefs and behavior produced.  This is like a version of
the abstraction and explanation capability that I have hypothesized.
One possibility is that the interpreter module \emph{is} this
hypothesized capability; another is that it evolved separately---but
it is difficult to see its utility prior to shared intentionality, and
it would be redundant afterward.  There are opportunities for
further investigation here.

\section{Comparison with Other Theories of Consciousness}

There are many theories of consciousness (of which I focus on the
materialist variety): some such as Neurobiological Naturalism (NN)
posit ancient evolutionary origins \cite{Feinberg&Mallatt16}; others,
such as Global Workspace Theories (GWT) \cite{Baars05:GWT,Dehaene14},
focus on biological processes; some, such as Integrated Information
Theory (IIT) \cite{Tononi-etal16:PSC}, Orch-OR
\cite{Hameroff&Penrose13}, and Panpsychism \cite{SEP:Panpsychism},
favor physical explanations and mechanisms; yet others, such as Higher
Order Thought (HOT) \cite{Gennaro04,Rosenthal04,Brown-etal19} and
Attention Schema Theory (AST) \cite{Graziano13} hypothesize
architectural ``dual-process'' structures in the brain
\cite{Frankish10,Evans&Stanovich13,Kahneman11}.  Graziano
\emph{et al.} \cite{Graziano-etal20} compare and reconcile several of
these theories with AST, and the comparisons remain largely valid with
SIFT substituted for AST\@.  However, none of these other theories
claim to explain what consciousness---and phenomenal consciousness in
particular---\emph{does}, nor what it is \emph{for}.

SIFT is different in that it focuses on specific purposes to be
accomplished, and develops mechanisms to achieve these, starting with
shared intentionality (to achieve teamwork) and proceeding to
rationality, which is seen as a fortuitous side-effect, built on
shared intentionality: ``teamwork for one''.\footnote{For example, at
some time, we have surely all said ``I cannot explain it, but I can
show you how to do it.''  Elsewhere, we suggest how a task description
can be inferred from demonstrations by inverse reinforcement learning
\cite{Jha&Rushby19}; thus, Alice can construct an abstract model of
some task that her Lower System ``knows'' how to do by mentally
demonstrating it to herself.}  Intentional consciousness is then
identified with awareness of attention to, and control of, the
mechanisms of shared intentionality and its Upper State and models.
Phenomenal consciousness arises because we need to communicate aspects
of our sense experience and subconscious Lower State: hence these must
be abstracted into the Upper State of shared intentionality, of which
we are conscious.

My presentation of proposed mechanisms provides context for several of
the theories identified above; in particular, it delivers a ``dual
process'' architecture that is consistent with, and explains some
aspects of, existing dual-process models of brain function
\cite{Frankish10,Evans&Stanovich13,Kahneman11}.  Dual-process models
hypothesize a logical, not physical, organization of the brain, and it
is quite possible that they are realized by physical and neuronal
mechanisms with quite a different structure.  Plausible candidates
include global workspace theory \cite{Baars05:GWT} and global neuronal
workspace \cite{Dehaene14}.

The Upper System in my dual-process model, performs calculations whose
inputs and outputs are subconscious mental states located in the Lower
System: it is a part of the brain that senses and writes to other
parts of the brain.  While the subconscious Lower System builds
representations and models of the external world, the Upper System
builds representations of those representations; these constitute what
computer scientists call a ``reflective system''
\cite{reflection-wiki} and what philosophers refer to as
``Higher-Order Thought'': that is, thoughts about thoughts
\cite{Gennaro04,Rosenthal05}.\footnote{This is among the oldest
conceptions of consciousness, dating back at least to Locke in
1689: ``Consciousness is the perception of what passes in a man's own
mind'' \cite[Book II, Chapter 1, Section 19]{Locke:understanding}.}

Dual-process and HOT theories generally hypothesize some process
whereby activity in the subconscious results in awareness at an upper
or higher-order level but do not describe what purpose this serves.
Unlike most of these theories, SIFT
explains what the upper system does (and hence why it might have
evolved).  Furthermore, as Graziano observes \cite{Graziano13},
higher-order awareness must also affect the subconscious lower-level
activity (otherwise it is impotent), and few theories address this.
In SIFT, these two directions respectively correspond to
abstraction and concretion of explanations that are mechanized by the
well-accepted operations of predictive processing, and both are
involved and coupled in the construction of upper level models and
explanations, and lower level behavior.\footnote{Oddly, most HOT
theories do not relate their higher-order aspect to dual-process
theories, nor do they invoke PP as a mechanism that can build the
higher-order system; an exception is Lau \cite{Lau:consciousness22},
who does mention PP but opts for a first-order theory.}

Graziano's AST \cite{Graziano13} and some HOT theories
\cite{Brown-etal19} are the ones closest to SIFT: in our terminology,
Graziano proposes that the brain constructs a model (he calls it a
schema) of the targets of attention and it is ``aware'' of this model
and that awareness constitutes intentional
consciousness.\footnote{``Awareness is a schematic, informational
model of something, and attention is the thing being modeled''
\cite{Graziano:speculations14}.}  SIFT proposes that our Upper System
builds models of the Lower System and these support the construction
of explanations; intentional consciousness then corresponds to
awareness of the single-threaded process of attention and control that
manages and applies the resources of the Upper System to these tasks.
Graziano has to postulate the attention schema because he does not
otherwise have an upper-level system that can provide a location for
consciousness, whereas SIFT has its Upper System, so that
consciousness can reside directly with its processes of attention and
control.

This implies that consciousness in SIFT applies to a specific locus of
attention and control---that concerning the Upper System and its
construction of Upper Models and explanations---whereas AST applies it
more generally.  For example, Graziano believes
that higher animals are aware (i.e., conscious) of other animals'
focus of attention (e.g., ``he is looking at me'')
\cite{Graziano19:others}, whereas I consider this level of (mutual)
attention could be unconscious and would become conscious only if
elevated into an Upper System and formulated as the \emph{explicit}
explanation or thought ``he is looking at me.''

Consciousness in SIFT is awareness of attention and control in the
Upper System; some experiments are claimed to demonstrate that
attention and awareness are different
\cite{Lamme:attention-awareness03} and likewise attention and
consciousness \cite{Koch:attention-consciousness07}.  However, when I
say ``consciousness is awareness of attention\ldots,'' I am merely aligning
with Graziano's usage and do not intend a specific interpretation of
``awareness'' distinct from consciousness; I could equally well have
said ``consciousness \emph{is associated with} attention\ldots''
Furthermore, as noted above, this association is not with attention
generally, but with attention to the Upper System and its construction
of Upper Models and explanations.  Thus, I interpret the experimental
findings as applying to attention in the Lower System and not to that
which I associate with consciousness.

SIFT delivers a very specific aspect of shared intentionality: how to
communicate a goal, plan, or idea from one individual to another,
prior to the evolution of language. (As noted earlier, it is difficult
to see how language could have evolved prior to shared intentionality
but plausible that it could do so afterward, given the mechanisms
described here.)  However, there are precursors to shared
intentionality that provide related capabilities; these include the
``social brain'' (i.e., living in groups with complex social systems)
\cite{Dunbar98}, ``cooperative communication'' where individuals
``align their mental states with respect to events in their shared
environment'' \cite{Vasil-etal20}, ``shared agency'' \cite{Bratman13},
which describes the mutual plans that shared intentionality needs to
bring about in order to achieve teamwork, and the general ``theory of
mind'' \cite{iep-theory-of-mind}.  These provide some of the necessary
milieu for the emergence of shared intentionality as considered here,
but do not substitute for it.  I should also note that we have focused
here on construction of a single communication within a process to
achieve shared intentionality on some topic.  It may take more than
one communication to convey a complete idea and Jha and I suggest how
inverse reinforcement learning could be used to accomplish this over a
series of communications \cite{Jha&Rushby19}.

There is substantial and significant prior work on shared
intentionality by Tomasello and others
\cite{Tomasello:origins-communication10,Tomasello:origins-cognition08,Tomasello&Carpenter07},
but this work tends to focus on what we might call ``immediate''
intentionality, founded on joint attention, such as sharing
information (``there is food over there'') and goals (``let's go to
the water hole''), and not the communication of new ideas and future
plans.  I suggest that shared immediate intentionality can build
cooperation, but not teamwork, and it is teamwork that truly sets
humans apart.  Similarly, work on the origins of language suggests
several sources, such as child rearing \cite{Falk:toungues09} or
self-advertising \cite{Dessalles:why-talk07} but not shared
intentionality---with the exception of Bickerton
\cite{Bickerton:Adam09} who relates language with the organization of
scavenging, which I would again classify as shared immediate
intentionality.

I have not found prior work that explores mechanisms for shared
intentionality of the kind considered here, nor any that derive
rationality and consciousness from shared intentionality, but there is
work that relates aspects of consciousness to group communication.
Frith, in an influential paper of only two pages \cite{Frith95},
``sketches a conjecture'' that consciousness enables interaction with
others: ``Shareable knowledge (which I equate with the contents of
consciousness) is the necessary basis for the development of language
and communication.  In this account, the major mistake of most
theories of consciousness is to try to develop an explanation in terms
of an isolated organism.''

Oakley and Halligan \cite{Oakley&Halligan17} give an account similar
to Frith's, but with more detail; they claim that consciousness
has no executive function and is basically a ``personal narrative''
about processes and actions generated by nonconscious systems.
Aspects of this narrative can be shared with others through ``external
broadcasting'' and this provides evolutionary benefits.  They discuss
the experimental literature and cite several others (including some
mentioned here) who ``accept that any evolutionary advantage lies not
in the `experience of consciousness' itself, but in the ability of
individuals to convey selected aspects of their private thoughts,
beliefs, experiences etc.\ to others of their species.''

Along these lines, Baumeister, Masicampo, and DeWall claim that ``the
purpose of human conscious thought is participation in social and
cultural groups'' \cite{Baumeister&Masicampo10}.  They see reasoning
and intentional consciousness as serving higher-level purposes that
make groups more effective but do not single out shared
intentionality.  Humphrey also associates phenomenal consciousness (he
calls it ``sentience'') with social purposes
\cite{Humphrey:sentience22}.  Similarly, de Bruin and Michael
\cite{deBruin&Michael18} suggest that Predictive Processing with upper
level models informed by a theory of mind enables effective social
cognition, while Sperber and Mercier \cite{Mercier&Sperber11} posit
that the purpose of human reasoning is evaluation of possibly false
information supplied by others; Dessalles \cite{Dessalles:why-talk07}
attributes similar functions to language.

All these authors accept or assume that (what I characterize as)
shared intentionality is among the collection of capabilities
associated with consciousness and that the collection provides
evolutionary benefit.  However, they seem, implicitly, to assume
``consciousness first'' or ``reasoning first'' theories rather than
giving primacy to shared intentionality, and they lack models for the
underlying representations, algorithms, and implementations.  They
assume an ability to manipulate and communicate concepts without
proposing how this can be constructed on more basic foundations, and
do not provide a strong path from one capability to another, nor do
they explain phenomenal consciousness.

In contrast, I propose that ``Shared Intentionality First'' provides
the most plausible basis for the construction of the package of
capabilities that includes consciousness and rationality, and that
consideration of the mechanisms, representations, and algorithms
required for its construction leads quite naturally to the other
components of the package, including phenomenal consciousness.

\section{Artificial Consciousness}
\label{artconsc}

There is a maxim, generally attributed to Richard Feynman, that to
really understand something you have to be able to recreate
it.\footnote{Written on his blackboard at the time of his death: ``What I
cannot create I do not understand.''}  Accordingly, there is a
subfield of consciousness research that explores the possibility of
building conscious robots, typically based on some theory of human
consciousness.  At the very least, these endeavors force elaboration
of sufficient detail in the chosen theory that it can be simulated in
a computational agent, and they also force articulation of what
consciousness might be in such agents and how it can be detected.

We could apply this to the SIFT hypothesis and ask whether robots
constructed along the lines described for Alice and Bob might be
conscious---and if not, we could ask whether this casts doubt on the
hypothesis.  Before doing this, we review something of the history of
attempts to develop conscious robots, also referred to as artificial
or machine consciousness.  This discussion is based on \cite[Section
5]{Rushby&Sanchez18}.

\exmemo{Angel needs better description; don't say agent.

In 1989, Leonard Angel published a book with the provocative title
``How to Build a Conscious Machine'' \cite{Angel89} in which he
proposed a kind of agent system.}

Early experiments were conducted by Tiham\'{e}r Nemes in the 1960s
\cite{Nemes70}, but intelligence and consciousness were not sharply
distinguished at that time, nor were cybernetics and (what became)
AI\@.  A modern view of robot or artificial consciousness is
attributed to Igor Aleksander in 1992 \cite{Aleksander92}, who
postulated that such a robot would need representations for depiction,
imagination, attention, planning, and emotion, and that consciousness
could emerge from their interaction.

The first large project to explore artificial consciousness was {\sc
cronus} \cite{Marques-etal07}.  This was predicated on the idea that
internal models of the system's own operation (i.e., what computer
scientists call ``reflection'' \cite{reflection-wiki}, and the related
philosophical notion of a ``self-model'' \cite{Metzinger:self-models07})
play an important part in consciousness.  Physically, {\sc cronus} was
an \emph{anthropomimetic} robot (i.e., one closely based on the human
musculoskeletal system) equipped with a soft-realtime physics-based
simulation of itself in its environment.  The internal simulation
allowed the robot to project the effects of possible future actions,
which the authors describe as ``functional imagination''
\cite{Marques-etal08}.  Later studies used an even more complex robot
(``{\sc eccerobot}''), while earlier ones had used a very simple,
nonanthropomorphic device \cite{Holland03}.  It is debatable whether
complex robots added a great deal to these experiments, and they
certainly increased the engineering challenges.

Like {\sc cronus}, most recent explorations of artificial
consciousness generally favor reflective architectures that employ
explicit models of self.  Experiments by Chella and colleagues
explored such robots' interaction with others
\cite{Chella&Manzotti09,Chella-etal08}; here, models of self applied
to ``others like me'' provide a theory of mind \cite{Winfield18:ToM},
and scenarios enacted by these models can be communicated to others
(directly, not in the manner proposed for Alice and Bob) to create a
form of shared intentionality \cite{Winfield18:ToM}.  These
capabilities can be used for ``inner dialog'' that provides
rationality in a way that resembles the local loop of Section
\ref{rationality} \cite{Pipitone-etal19,Holland03};

Gamez describes other projects performed around the same time
\cite{Gamez08}.  All these experiments, and those mentioned above,
employ some form of reflection or HOT as their underlying theory of
consciousness.  Others have built systems based on GWT or IIT; Reggia
provides a survey \cite{Reggia13}.  None of these projects, nor those
mentioned earlier, claim to have demonstrated artificial consciousness
and I suspect the same would be true of Alice and Bob, despite their
design being based on mechanisms thought to correspond to those of
humans.

Research on artificial consciousness seems not to have a central forum
for presentation of results and discussion of ideas: the
\emph{International Journal of Machine Consciousness} began
publication in 2009 but ceased in 2014.\footnote{It has recently been
revived as the Journal of Artificial Intelligence and Consciousness.}
Perhaps as a result, recent work seems to retread familiar ground.
For example, a paper by Dehaene, Lau and Kouider from 2017
\cite{Dehaene-etal17} presents the authors' theory of consciousness
(global availability as in GWT, plus reflection built on PP), then
asserts that a machine with these capabilities ``would behave as
though it were conscious'' \cite{Dehaene-etal17}.  In a response,
Carter \emph{et al.}\ \cite{Carter-etal18} observe that Dehaene and
colleagues ask and answer the wrong questions---essentially, Dehaene
\emph{et al.}\ are aiming for intentional consciousness, whereas
Carter \emph{et al.}\ think that phenomenal consciousness is what
matters: for machines to be conscious, ``we must ask whether they have
subjective experiences: do machines consciously perceive and sense
colors, sounds, and smells?''  They posit ``a more pertinent question
for the field might be: what would constitute successful demonstration
of artificial consciousness?''  This is an old question (e.g.,
\cite{Boltuc09}) that still seems to lack good answers.

An event in June of 2022 illustrates this: a Google engineer working
with their Large Language Model (LLM), LaMDA (Language Model for
Dialogue Applications), claimed it had become ``sentient'' and
possibly conscious (Washington Post, 11 June 2022).  A flurry of
discussion ensued, with most commentators rejecting the claim of
consciousness, but lacking a firm basis for doing so
\cite{Shardlow&Przbyla22}.  Modern LLMs easily pass standardized tests
for intelligence and knowledge (e.g., high school, medical, and law
examinations) and, arguably, traditional benchmarks for human-level
cognition such as the Turing Test \cite{Turing:test50} (thereby
motivating proposals for more demanding tests
\cite{Shrivastava-etal:imitation22}).  Furthermore, they can generate
interactive text that speaks coherently about feelings and
experiences.  On the other hand, there is nothing in their internal
operation that resembles any theory of consciousness, and no credible
explanation how consciousness might emerge from the way they do
operate.

Although deliberate and accidental research has not unequivocally
demonstrated artificial consciousness, it is sharpening discussion on
how consciousness could be detected, particularly since it might not
resemble human consciousness (the conjecture of octopus consciousness
raises the same issue).  Some of these discussions distinguish
\emph{strong} and \emph{weak} forms of artificial consciousness
\cite{Holland:machine-consciousness03} (sometimes framed as
duplication vs.\ simulation).  Strong artificial consciousness would
\emph{be} conscious, whereas the weak form exhibits behaviors and
attributes associated with consciousness without actually possessing
it (cf.\ ``philosophical zombies,'' which are considered below).

Most researchers think that simulations of theories of human
consciousness can create, at best, the weak form of artificial
consciousness, and that the weak form does not lead to the strong.  By
analogy, we can build extremely accurate simulations of the cosmos and
explore the missing mass (attributed to dark matter and dark energy),
yet the simulations do not \emph{have} mass; so a simulation \emph{of}
consciousness will not \emph{have} consciousness.

On the other hand, the weak and strong distinction seems to matter
only for phenomenal consciousness: we likely will regard an entity
that \emph{has} feelings differently than one that merely simulates
them.  But weak intentional consciousness is operationally equivalent
to the strong form: if weak intentional consciousness enables some new
cognitive capabilities, then the underlying system can strongly
possess these by running the weak simulation as a subroutine.  This
asymmetry between weak and strong forms of phenomenal and
intentional consciousness is related to the ``hard problem'' of
consciousness \cite{Chalmers95:hard-problem} and provides
another way to formulate it.

These observations have some impact on the possibility of
philosophical zombies, which are hypothetical entities built on a
biological substrate that lack consciousness but reproduce levels of
cognitive and social performance that are indistinguishable from
conscious humans \cite{sep-zombies}.  The question is whether such
entities are possible.  As we have postulated them, biological zombies
lacking intentional consciousness would be unable to focus and control
the operation of their Upper System and would be unable to communicate
the content of their sense experience; they would be unable to utter
truthfully such simple phrases as ``I smell a rat.''  However, we
could postulate entities with computational mechanisms that resemble
our humanoid robots and thereby possess the behavioral characteristics
of intentional consciousness without the experience of phenomenal
consciousness.  If such entities chose to fake the experience (as
modern LLMs would enable them to do), they would come very close to
philosophical zombies.

The general lack of success in efforts to create artificial
consciousness, and its likely absence in the Alice and Bob robots, may seem to
cast doubt on the mechanisms hypothesized here to create consciousness
in humans.  My opinion is that these judgements are premature: just as
we have machines that fly but do not flap their wings, and we had to
acquire deep knowledge of aerodynamics to reconcile these different
approaches to flight, so our characterizations of consciousness may be too crude
and coarse to identify significant nascent properties in Alice and
Bob.  Birch \emph{et al.} nominate five separate elements in
consciousness that should be investigated individually
\cite{Birch-etal20}, while Holland suggests that artificial
consciousness should be reframed independently of organic life forms
\cite{Holland:bat20}.

\section{Summary and Conclusions}

The theory and speculation I have advanced here as ``SIFT'' comprise
several basic claims that are somewhat independent and can be
evaluated separately.  The initial and central claim is ``Shared
Intentionality First'': this is the idea that human rationality and
consciousness emerge from a framework that constructs shared
intentionality prior to the evolution of language and that,
altogether, these faculties constitute a ``package'' of capabilities.
This claim is independent of the mechanism and biological
implementation of that framework.  Shared intentionality enables
teamwork, which confers evolutionary advantage.  Given shared
intentionality, an individual can use it locally (i.e., communicate
with herself) and that provides rationality; consciousness is
awareness and control of these faculties in operation, and phenomenal
consciousness arises so that sense experience can be communicated.

Second is the claim that shared intentionality requires mechanisms for
abstraction, concretion, and explanation, and therefore Figure
\ref{modelpic1} is correct: there simply has to be a mechanism that
abstracts part of the subconscious neural state into a succinct
representation based on shared concepts and arranged as an explanation
that can be communicated to others.  An inverse mechanism, concretion,
translates the explanation back into internal mental states that
enrich the receiver's subconscious so that he can now deliver usefully
cooperative behavior.  The communicated explanation is not just a
collection of concepts, it should employ structures to indicate logical
operations (``and,'' ``or''), temporal sequencing (``before,''
``then''), causation (``because,'' ``in order to'') etc., and the
mechanisms for abstraction and concretion must be able to construct,
deconstruct, and otherwise manipulate these.  Explanations are
wordless and communicated by mime, demonstration, or symbolic
gestures and sounds, but I speculate they could provide a foundation for
``mentalese'' and, ultimately, language and speech.

Third is the claim that, to be effective, the explanation must be
constructed using a model of the receiver's concretion operation: that
is, a theory of mind with estimates of the receiver's
knowledge and beliefs.  To mechanize this, the sender can use suitable
adjustments to her own concretion function, so that abstraction and
generation of the explanation are achieved by an optimizing search to
find an explanation whose adjusted concretion matches elements of the
sender's own subconscious state.  This argues that concretion and
abstraction are closely associated and likely reside in the same
mental unit, as portrayed in Figure \ref{modelpic2}.  Predictive
Processing is a biological process that can achieve these algorithmic
requirements, whereby concretion and abstraction are represented by
prediction and prediction error, respectively, and approximations to
Variational Bayes use these to optimize abstract models and
explanations.  It is not necessary that these computational processes
comprise or are precursors to a dual-process logical architecture, but it
does seem plausible.

Fourth is the claim that local operation of the concretion/abstraction
loop provides deliberation and rationality.  Components of the loop
have the ability to reason on explanations (as just described), and
can recruit automated subconscious capabilities as subroutines via
metaphorical translations (e.g., to reason about social hierarchy via
translation to automated capabilities for understanding vertical
space).  Manipulation of the concretion operation (via its
parameterization by a theory of mind) allows counterfactual and
hypothetical reasoning.  In their basic form, these capabilities
``fall out'' of the mechanisms for shared intentionality but may be
enhanced with more sophisticated control, as portrayed in Figure
\ref{solo}.

Fifth is the speculative claim that awareness of attention to
explanations and their associated models, together with control of
their generation and interpretation, constitutes intentional
consciousness.  The claim explains the underlying purpose and
construction of intentional consciousness but does not explain how or
why it delivers a subjective experience: that aspect is speculative.
I do not know of a way to confirm or refute this speculation but note
that the underlying construction does explain otherwise puzzling
facts, such as why the conscious mind is less an initiator of actions
and more a reporter and interpreter of actions and decisions initiated
elsewhere in the brain, why much of the experience of consciousness is
of an inner dialog (rather than, say, a stream of images), and why the
experience is at an intermediate level of representation.  Thus,
although the combination of Claims 1 to 5 explains the purpose of the
Upper State and its models and the need for control and attention to
its processes, I accept that they do not explain how these deliver
consciousness as a nonphysical experience.  (I do not apply to SIFT
the illusionist step that some use to equate a report
of consciousness with consciousness itself, although others are free
to do so.)

Sixth is the claim that phenomenal consciousness is required so that
we can communicate our sense experiences: ``I just heard the roar of a
lion.''  The content of some of our senses and subjective sensations
(e.g., pains and emotions and moods) must be available for abstraction
and incorporation into explanations for communication to others.
Representations of these qualia are delivered to the
abstraction/explanation mechanism where conscious resides, and
therefore become conscious.  The representations used for qualia are
unimportant as long as they can be distinguished from one another, but
we care about the representations actually used and refer to their
experience as phenomenal consciousness.

These six claims are different from most other explanations and theories
of consciousness and related topics in that they focus on functions
needed to achieve certain purposes, and on mechanisms to deliver them.
However, they are not incompatible with other theories and can be seen
to provide context for several of them.

In particular, the combination of an abstraction function going ``up''
and a concretion function going ``down,'' engaged together in an
optimizing search to construct upper level models and communicable
explanations, is consistent with predictive processing models of brain
function \cite{Clark13,Hohwy:book13,Friston:free-energy10}.  Thus, the
biological plausibility of predictive processing and exploration of
its neural correlates \cite{Hohwy&Seth20} can also support my
claims (although these observations about PP are not unchallenged
\cite{Schlicht&Dolegna22}).

Similarly, my presentation of proposed mechanisms derives a ``dual
process'' architecture that is consistent with, and explains some aspects of,
existing dual-process models of brain function
\cite{Frankish10,Evans&Stanovich13,Kahneman11}.  Dual-process models
hypothesize a logical, not physical, organization of the brain, and it
is quite possible that they are realized by physical and neuronal
mechanisms with quite a different structure, as hypothesized by global
workspace theories \cite{Baars05:GWT,Dehaene14}.

The Upper System in my dual-process model, performs calculations whose
inputs and outputs are subconscious mental states: it is a part of the
brain that senses and writes to other parts of the brain.  This is
consistent with Higher-Order Thought
\cite{Gennaro04,Rosenthal05,Brown-etal19} and, when control of these
processes is associated with attention, to Attention Schema Theory
\cite{Graziano13}.

The six claims for SIFT are related and cumulative but differ in their
scientific positioning.  The fundamental criterion for a scientific
theory is that it is testable and, moreover, falsifiable
\cite{Popper:2014logic}.  Schurger and Graziano make a further
distinction: between scientific \emph{laws} and \emph{theories}
\cite{Schurger&Graziano22}; both can be falsifiable but laws provide
only descriptions (e.g., Newton's law of gravitation), whereas
theories deliver explanations (e.g., General Relativity).  They find
that most theories of consciousness are, at best, laws and that only
AST qualifies as a theory (although they concede it is a theory for
\emph{belief} in consciousness, rather than for consciousness itself).

Considering our six claims, the first (i.e., shared intentionality is
the foundation for rationality and consciousness) is a testable and
falsifiable theory about human evolution.  The second and third
(mechanisms for shared intentionality) are verifiable computer science
constructions that deliver a theory about human cognition that seems
testable by methods from psychology.  The fourth (rationality) is a
speculative theory, but it seems testable in both robots and humans.
The fifth provides an explanation for the function and construction of
the processes that underlie intentional consciousness.  Their
testability awaits development of more refined criteria for evaluating
consciousness.  However, this theory does not explain how or why
intentional consciousness produces a subjective experience.  The sixth
claim is a logical consequence of the other five and delivers a strong
explanation for the purpose and construction of phenomenal
consciousness.  Thus, SIFT fares at least as well as any description
under Schurger and Graziano's criteria, and provides explanations that
are both more comprehensive and more specific than other theories.

In conclusion, I propose that the Theory of ``Shared Intentionality
First'' (SIFT) provides the most plausible basis for development of
the cognitive package that includes consciousness and rationality.
The theory explains the functions performed by intentional and
phenomenal consciousness and how they are constructed.  It reduces the
``hard problem'' for phenomenal consciousness to that for intentional
consciousness but, like all other theories of consciousness, it is
unable to explain why the latter produces an apparently nonphysical
experience.  I invite others to develop or contest these proposals.

\subsection*{Acknowledgments}

I am grateful to Antonio Chella and Owen Holland for advice and
guidance to the literature.  Owen Holland  provided invaluable
advice on previous work and relevant topics, and on framing these
ideas for an audience beyond computer science.  Harold Thimbleby
alerted me to David Marr's work.  

Colleagues at SRI, particularly Maria Paola Bonacina, Susmit Jha,
Prashanth Mundkur, and N. Shankar have been patient and critical
sounding boards.  Prashanth provided extensive and helpful commentary.
I greatly appreciate the support and encouragement of my boss, Pat
Lincoln.

\vspace*{-1.5ex}
\section*{References}
\addcontentsline{toc}{section}{References}
\vspace*{-1.5ex}
\bibliographystyle{modplain}

\end{document}